%% file: main.tex
\documentclass[sigconf, anonymous=false]{acmart}
  \providecommand\BibTeX{{%
    \normalfont B\kern-0.5em{\scshape i\kern-0.25em b}\kern-0.8em\TeX}}

\usepackage{graphicx}

\usepackage{dsfont}
\usepackage{algorithm}
\usepackage{algorithmic}
\usepackage{xspace}
\usepackage{enumitem}

\newtheorem{proposition}{Proposition}
\newtheorem{lemma}{Lemma}

\newtheorem{definition}{Definition}

\newcommand{\thesystem}{\textsc{WebGuard}\xspace}

\definecolor{ForestGreen}{rgb}{0.0, 0.65, 0.31}

\newcommand{\bo}[1]{{\color{orange}$<$} \textbf{\textit{BO: }}{\color{orange} #1} {\color{orange}$>$}}

\begin{document}

\pagenumbering{arabic}
\thispagestyle{plain}
\pagestyle{plain}

%%
%% The "title" command has an optional parameter,
%% allowing the author to define a "short title" to be used in page headers.
\title{In-Application Defense Against Evasive Web Scans through Behavioral Analysis}

%%
%% The "author" command and its associated commands are used to define
%% the authors and their affiliations.
%% Of note is the shared affiliation of the first two authors, and the
%% "authornote" and "authornotemark" commands
%% used to denote shared contribution to the research.

\author{Behzad Ousat}
\affiliation{%
  \institution{Florida International University}
  % \city{New York}
  % \country{USA}
  }
\email{bousat@fiu.edu}

\author{Mahshad Shariatnasab}
\affiliation{%
  \institution{Florida International University}
  % \city{New York}
  % \country{USA}
  }
\email{mshar075@fiu.edu}

\author{Esteban Schafir}
\affiliation{%
  \institution{Florida International University}
  % \city{New York}
  % \country{USA}
  }
\email{escha032@fiu.edu}

\author{Farhad Shirani Chaharsooghi}
\affiliation{%
  \institution{Florida International University}
  % \city{New York}
  % \country{USA}
  }
\email{fshirani@fiu.edu}

\author{Amin Kharraz}
\affiliation{%
  \institution{Florida International University}
  % \city{New York}
  % \country{USA}
  }
\email{ak@cs.fiu.edu}

% }
%%
%% By default, the full list of authors will be used in the page
%% headers. Often, this list is too long, and will overlap
%% other information printed in the page headers. This command allows
%% the author to define a more concise list
%% of authors' names for this purpose.
\renewcommand{\shortauthors}{Behzad Ousat, et al.}

\input{00-Abstract}

%%
%% The code below is generated by the tool at http://dl.acm.org/ccs.cfm.
%% Please copy and paste the code instead of the example below.
%%


%%
%% Keywords. The author(s) should pick words that accurately describe
%% the work being presented. Separate the keywords with commas.

%% A "teaser" image appears between the author and affiliation
%% information and the body of the document, and typically spans the
%% page.
% \begin{teaserfigure}
%   \includegraphics[width=\textwidth]{sampleteaser}
%   \caption{Seattle Mariners at Spring Training, 2010.}
%   \Description{Enjoying the baseball game from the third-base
%   seats. Ichiro Suzuki preparing to bat.}
%   \label{fig:teaser}
% \end{teaserfigure}

% \received{20 February 2007}
% \received[revised]{12 March 2009}
% \received[accepted]{5 June 2009}

%%
%% This command processes the author and affiliation and title
%% information and builds the first part of the formatted document.
\maketitle

\input{01-Introduction}
\input{02-Background}

\input{03-Methodology}

\input{04-Analysis}

\input{05-Results}
\input{06-Discussion}

\input{07-Conclusion}

\bibliographystyle{ACM-Reference-Format}
\bibliography{references}

%%
%% If your work has an appendix, this is the place to put it.
% \appendix

\input{09-Apendix}

\end{document}

%% file: 00-Abstract.tex
%-------------------------------------------------------------------------------
\begin{abstract}
%-------------------------------------------------------------------------------
%Web scanners are automated software systems designed to interact with web applications.
Web traffic has evolved to include both human users and automated agents, ranging from benign web crawlers to adversarial scanners such as those capable of credential stuffing, command injection, and account hijacking at the web scale. 
%In recent years, these operations have become highly effective in exhausting defensive resources or finding vulnerabilities in modern web applications.
%Today, these scanners are responsible for at least 20\% of web traffic
The estimated financial costs of these adversarial activities are estimated to exceed tens of billions of dollars in 2023.
In this work, we introduce \thesystem, a low-overhead in-application
forensics engine,
to enable robust identification and monitoring of automated web scanners, and help mitigate the associated security risks. \thesystem
%to force a robust tradeoff between achieving more visibility and increasing the cost of evasion. That is, \thesystem 
 focuses on 
%This has led to an urgent need to develop efficient bot detection mechanisms 
%satisfying 
the following design criteria: (i) integration into web applications without any changes to the underlying software components or infrastructure, (ii) minimal communication overhead, (iii) capability for real-time detection, e.g., within hundreds of milliseconds, and (iv) attribution capability to identify new behavioral patterns and detect emerging agent categories.
To this end, we have equipped \thesystem with multi-modal behavioral monitoring mechanisms, such as monitoring spatio-temporal data and browser events.
%A lightweight JavaScript library is introduced to enable integration of WebGuard into web applications via a script tag.  
%a wide range of run-time behavioral data 
%for remote analysis and response
We also design supervised and unsupervised learning architectures for real-time detection and offline attribution of human and automated agents, respectively. 
%These multi-modal artifacts are used for real-time detection and attribution achieving accuracy above 90\% in less than one second of monitoring the user, while introducing \textcolor{red}{X} FP.  
Information theoretic analysis and empirical evaluations are provided to show that multi-modal data analysis, as opposed to uni-modal analysis which relies solely on mouse movement dynamics, significantly improves time-to-detection and attribution accuracy. Various numerical evaluations using real-world data collected via \thesystem are provided achieving 
 high accuracy in hundreds of milliseconds, with a communication overhead below 10 KB per second.

% Theoretical analysis and empirical evaluations are provided to show the effectiveness of multi-modal approach in increasing the accuracy of attribution mechanism and real-time detection based on several experiments conducted in a real-world setting. 

%show that multi-modal data collection, as opposed to uni-modal approaches, %which relies solely on mouse movement dynamics,
%significantly improves detection time and clustering accuracy. Various numerical evaluations of the real-world data
%collected via WebGuard are provided to provide additional
%insights.

%On the client side, a lightweight JavaScript library is employed that can be embedded into web applications via a script tag, addressing criterion (i). Event listeners are used to capture a rich set of behavioral data, such as mouse movements, clicks, scrolls, and clipboard actions. Empirical analysis is provided to show a communication overhead of less than \textcolor{red}{10 Bytes/second?}, addressing  criterion (ii). Supervised and unsupervised learning architectures are introduced for real-time detection and offline clustering, respectively,  achieving accuracies above $\%95$ and satisfying criteria (iii) and (iv). Theoretical analysis and empirical evaluations are provided to show that multi-modal data collection, as opposed to uni-modal which relies solely on mouse movement dynamics, significantly improves detection time and clustering accuracy. Various numerical evaluations of the real-world data  collected via WebGuard are provided to provide additional insights.  
\end{abstract}

%% file: 01-Introduction.tex
\section{Introduction}
\label{sec:int}

Web traffic today is increasingly dominated by automated agents and web scanners, with the share of automated traffic almost equalling the share of human traffic as of 2022 \cite{imperva2023}. 
While agents such as search engine crawlers are essential for the web's functioning \cite{olston2010web}, there's a rising tide of malicious agents. These range from DDoS attackers aiming to disrupt services, to data scrapers extracting information without permission, and bots testing sites for security vulnerabilities \cite{mittal2023deep,khder2021web}. Their increasing volume and sophistication pose significant threats to online platforms and users alike. For instance, the Merchant Risk Council
(MRC), a non-profit association for payments and fraud prevention has reported that annual online fraud losses,
as a result of credential stuffing, are projected to exceed \$48 billion in 2023~\cite{isemag}.
%Detecting malicious automated agents in real-time, especially amidst legitimate web traffic, presents an urgent and critical challenge. 

To thwart the threat posed by malicious web scanners, many platforms employ CAPTCHAs (Completely Automated Public Turing test to tell Computers and Humans Apart) for bot detection \cite{von2003captcha}. CAPTCHAs are tests designed to be easily surmountable for humans but challenging for bots. 
For instance, previous versions of Google’s reCAPTCHA (v1 and
v2) present tasks involving images, letters, and audio, which are easily solved by humans but challenging for computers.
However, there are inherent limitations to these prior CAPTCHA versions: they are typically a one-time test, making them vulnerable to circumvention via crowd workers \cite{niu2023exploring}; advancements in machine learning techniques for image and text recognition have rendered even complex CAPTCHAs susceptible at minimal costs \cite{alqahtani2020image,stark2015captcha,george2017generative}; and frequent CAPTCHA prompts can be obtrusive, deteriorating the user experience. The text recognition tasks of reCAPTCHA v1 were broken by \cite{bursztein2014end}
achieving 98\% evasion rate. The visual and audio tasks of reCAPTCHA v2 were also broken using deep learning methods, e.g., using UnCAPTCHA techniques of \cite{bock2017uncaptcha}. 

In light of these limitations, there has been a pivot towards utilizing behavioral insights, such as mouse movement dynamics, for bot detection and authentication \cite{chong2019user,hamidzadeh2023frs,mondal2017study,acien2022becaptcha}. For example, in reCAPTCHA's third iteration,  Google ensures there's no disruption for users by running an adaptive risk analysis in the background. The goal of this updated system is to not just minimize interference in the user's experience, but also to obfuscate the nature of the challenge itself.
These passive mechanisms operate without impeding user activity and offer resilience against crowd worker attacks due to continuous monitoring. However, they may still be broken using sophisticated generative machine learning techniques capable of bypassing biometrics-based detections \cite{tsingenopoulos2022captcha}.
For example, \cite{akrout1903hacking} achieves a 97.4\% evasion success rate against reCAPTCHA v3 by using reinforcement learning techniques to generate mouse movement patterns.  
% Additionally, the time taken by these mechanisms to flag a malicious entity might be substantial, potentially allowing damage before detection.

 In this work, we introduce a multi-modal forensics engine, \thesystem, to address the aforementioned shortcomings in state-of-the-art web scanner detection systems. In general, for a detection system to be considered ideal, it must meet several key criteria. Firstly, it should easily integrate across a variety of web applications and platforms. Secondly, it must ensure that the user experience remains uninterrupted. 
 %and that the data collection does not breach user privacy. 
 Thirdly, the system must operate with minimal communication overhead to guarantee smooth operation. Fourthly, it should have the capability to detect web scanners in real-time, with a preferred response time within hundreds of milliseconds. Lastly, as malicious agents evolve, the system should be able to recognize new behavioral patterns, thereby enabling it to detect and combat emerging threats effectively. \thesystem satisfies the aforementioned criteria. It automatically sets event listeners for almost all of the forensically relevant interactions with the web application. Upon the activation of an event, it generates a callback that records both the temporal and spatial details of that event, indicating its timing and location on the page. The forensics engine can recognize and process 43 primary events supported by all major browser vendors. Additionally, it gathers supplementary metadata like the webpage's path, specific object IDs within the page's Document Object Model (DOM) ---including elements such as forms, text fields, passwords, checkboxes, and submit buttons --- and the respective actions taken.

Our analysis shows that the inclusion of multi-modal data significantly expands the capabilities of current uni-modal state-of-the-art systems. We illustrate these improvements both via empirical evaluation in Section \ref{sec:emp} and theoretical analysis in Section \ref{sec:th}. Furthermore, \thesystem can be incorporated into web applications via a lightweight Javascript tag which is applicable to both desktop and mobile users. We show through empirical analysis (Section \ref{sec:emp}) that the integration of the \thesystem engine induces minimal communication overhead, in the order of several KB per second.

 In addition to introducing the \thesystem forensics engine and its associated library, we develop an accompanying machine learning framework for real-time detection as well as an offline attribution mechanism~\footnote{ \url{https://github.com/multimodalforensics/webguard}}. In the attribution module, we wish to observe the website visitors for an extended
period of time (e.g., several days) and perform unsupervised learning on the collected data to identify different
visitor classes. 
\\The main contributions of this work are summarized below:
\begin{itemize}[leftmargin=*]
%\vspace{-.08in}
\item Present a multi-modal forensics engine, \thesystem, to bridge the gaps in state-of-the-art bot detection systems. (Section \ref{sec:thesystem})
%\vspace{-.08in}
\item Introduce an accompanying machine learning framework for real-time bot detection and offline attribution. (Section \ref{sec:ml})
%\vspace{-.08in}
\item Introduce a multi-class hypothesis testing framework that not only allows for accurate bot detection in less than one second but also distinguishes between various classes of malicious agents with high accuracy. (Section \ref{sec:ml})
%\vspace{-.08in}
\item Showcase thorough empirical evidence of the minimal communication overhead of \thesystem. (Section \ref{sec:emp})
\item Provide theoretical and empirical evaluations demonstrating the robustness of the multi-modal approach to generative and reinforcement learning-based attacks. In particular, it shows a linearly inverse relation between time to detection and the number of modalities. (Sections \ref{sec:emp} and \ref{sec:th})
%\vspace{-.08in}

\end{itemize}

%% file: 02-Background.tex
\section{Background}
\label{Background}
In this section, we provide a brief overview of the threat model used to emulate adversaries' capabilities in building various types of adversarial scanners. We describe some of the currently used defense mechanisms against adversarial scans,  
and discuss the background related to pattern recognition and machine learning techniques utilized in subsequent sections.

\subsection{Threat Model}
\label{sec:threat-model}
%Adversarial interactions with web applications can occur in different forms. Adversaries'
Malicious web scanners are often designed to locate and target web applications in a scalable and time-efficient way; hence automating
the necessary steps of identifying exposed services while evading common defense mechanisms. As a result, in our threat model, we make the following assumptions regarding the attacker's capabilities:
\begin{itemize}[leftmargin=*]
%\vspace{-.08in}
\item\textbf{Reporting Arbitrary Identities.} Adversaries can simulate regular web sessions by employing full-fledged browsers rather than using headless browsers or classic methods such as curl to satisfy common checks on the server;
%One of the adversaries' goals during the scan-abuse cycle is to automate 
%the necessary process of exploring the exposed services of web applications.
%The process often includes scanning the static structure of web applications, i.e., frames, forms, anchors, JavaScript events and record the sequence of steps required to reach a given %page. We assume that adversaries use any known techniques to build tools to simulate regular user session traffic during the scan-abuse cycle.
%\vspace{-.08in}
\item \textbf{Using Network Proxies.} The web traffic interacting with the target application can be generated from legitimate cloud infrastructure or compromised servers to reduce the effectiveness of common reputation-based techniques~\cite{10.1145-7,10.1145-2,10.5555,10.1145-3}
%,10.1007}
; 
%\vspace{-.08in}
\item \textbf{Simulating User Interactions.} The malicious code can interact with the target web application by clicking on elements or scrolling down the page to invoke a dynamic behavior of the target application. This allows adversaries to identify services that require dynamic interactions; 
%\vspace{-.08in}
\item \textbf{Using Artificial Delays.} The malicious code can insert artificial delays between every two consecutive requests to evade potential threshold-based or anomaly detection techniques that monitor sudden changes in the traffic behavior; 
%\vspace{-.08in}
\item \textbf{Passing Adversarial Inputs.}  Adversarial inputs and interactions can be generated --- for example, command injections and
adversarial fuzzing --- to influence the desired outcome or the security posture of the exposed service.
%\vspace{-.08in}
\end{itemize}
Our threat model considers the aforementioned risks. The solutions provided by \thesystem
are designed to be robust against such malicious activities.

\subsection{System Objectives}
There are various types of automated scanners each having different objectives such as content scraping, credential stuffing, DDoS, and account hijacking, among others. %content automation bots, and price scraping bots, among others.
Design of detection mechanisms which --- in addition to distinguishing between human and web scanners via binary hypothesis testing --- can distinguish among these different classes of scanners is of significant interest. The reason is that each class of web scanners has its own unique behavioral patterns, objectives, and operational tactics. A nuanced understanding and differentiation can lead to targeted defense mechanisms, allowing tailored countermeasures rather than generic responses. Moreover, this can provide insights into the intent and potential threat level, enabling a more proactive and granular security posture. Consequently, we pursue the following two main objectives in the design of \thesystem:
\begin{itemize}[leftmargin=*]
% %\vspace{-.08in}
\item \textbf{Real-time Detection.} The forensics engine is designed to equip web applications with the ability to both detect automated agents and categorize them into predefined classes. This dual functionality needs to occur in real-time, ideally within a time window of hundreds of milliseconds following the initiation of each session. The detection mechanism is made possible through the use of supervised learning techniques, trained on labeled data obtained from the offline attribution process described below.  
% %\vspace{-.08in}
\item \textbf{Offline Attribution and Trend Discovery.} The engine employs unsupervised machine learning algorithms to analyze extended sequences of behavioral data collected from unlabeled users. Through offline clustering, the engine is capable of identifying and grouping together users exhibiting similar behavioral patterns. By continuously monitoring the evolution of these clusters and their internal behavioral signatures, the engine allows clients to preemptively identify newly emerging trends in automated attacks. The output labels generated by this unsupervised learning phase serve as the ground truth for training supervised models used in the real-time detection mechanism, thus creating a feedback loop for continual model improvement.
\end{itemize}
\subsection{Classes of Automated Web Scanners}
\label{sec:bots}
  To illustrate the ability of \thesystem to detect and distinguish a specific type of automated agent among a set of given classes, we emulate the following scanner categories:
\begin{itemize}[leftmargin=*]
% %\vspace{-.08in}
    \item \textbf{Gremlins\cite{richard_2020}.} This user interface testing library is written in JavaScript. Gremlins is designed to test web applications by unleashing a horde of ``gremlins" to randomly execute functions and interact with elements on a page to simulate random user actions. These ``gremlin'' agents stochastically click on elements, fill forms, or move the mouse.
    % %\vspace{-.08in}
    \item \textbf{HLISA \cite{gossen2021hlisa}.}  Human-Like Interaction Selenium
API (HLISA)  is a library for simulating human-like interaction with web pages using Selenium.  HLISA mimics the speed, accuracy, and randomness of human mouse movement, clicks, scrolling, and typing. HLISA also spoofs browser properties that can be used for fingerprinting. We use the  HLISA library to emulate a crawler bot. 
%\vspace{-.08in}
\item \textbf{ZAP \cite{owasp_benchmark_2017}.} The Zed Attack Proxy (ZAP) is a widely used open-source web application security scanner developed by OWASP. It is designed to find vulnerabilities in web applications by simulating attacks. ZAP provides automated scanners as well as various other utilities. In the context of our work, ZAP can be considered as a tool that emulates advanced web application scanner bots.
%\vspace{-.08in}
\item \textbf{Random Bot Emulators.} In addition to the aforementioned libraries, we have designed and implemented two new random scanner classes, namely, Random Mouse Bot and Artificially Delayed Bot. These are developed using Puppeteer \cite{puppeteer} and each has different behavioral patterns. A more complete description is provided in Section \ref{sec:artifacts}.
\end{itemize}

\subsection{Prior Defensive Mechanisms}
In this section, we briefly discuss the practices in defense against adversarial scanning and automated attacks. 
\begin{itemize}[leftmargin=*]
% %\vspace{-.08in}
\item \textbf{Fingerprinting.} Conventional fingerprinting techniques~\cite{10.5555, NikiforakisKJKPV13, laperdrix2020browser, zhang2022survey} collect specific details about users such as hardware details, software details, and geolocation. Subsequently, they generate a fingerprint for that device. Fingerprints can be used for several purposes such as detecting previously observed adversarial agents attempting to visit the target web application. If an agent with automated activity is detected, the corresponding fingerprint is recorded for future checks. Fingerprinting mechanisms have been traditionally useful in tracking human users, however, they offer limited capabilities to identify modern evasive attacks~\cite{webbot,6575366,phishlab}. 
%In particular, one may generate malicious code that can report any arbitrary identity or establish connections from different parts of the web while launching attacks in a stealthy way -- as mentioned in the threat model. 
For instance, one method is to incorporate various network proxies and use less suspicious traffic sources and full-fledged browsers instead of using curl or headless browsers, and report arbitrary information while interacting with the web application to evade fingerprinting defenses.
%\vspace{-.08in}
\item \textbf{Traffic Attribution.} An emerging trend in detecting automated attacks is the rise of run-time analysis behavioral metrics.
For instance, DataDome~\cite{datadome}, Radware~\cite{radware}, Imperva\cite{imperva}, Baracuda Networks~\cite{baracuda} have introduced defenses focusing on attributes of incoming traffic to web applications. Features such as spikes in pageviews, high bounce rate (i.e., visiting a single page without clicking anything on the page), and low session duration due to automated clicking through pages have been used in current solutions~\cite{datadome1,cloudflare2} and are indeed %Incorporating these heuristics in the defense side is indeed
a good starting point to better analyze incoming web traffic.
 Recent bot detection systems have considered behavioral biometrics such as mouse movement dynamics along with metadata and activity logs\cite{chong2019user,hamidzadeh2023frs,mondal2017study,acien2022becaptcha}. 
Although these techniques can detect \emph{aggressive} scanners, they are not informative about the type of scanner and the behavioral statistics of evasive scanners that operate in a targeted and non-aggressive fashion, discussed in Section ~\ref{sec:threat-model}. In fact, in the most complex and consequential situations, evasive scanners may attempt to bypass these defense mechanisms by imitating basic human interactions, injecting artificial delays between consecutive requests, or artificially increasing the session duration by automatically triggering specific events periodically. 
Furthermore, the diversity of new forms of web attacks is expected to significantly increase as generative models advance in imitating human interaction, making it increasingly difficult to analyze the intent of remote agents. 
%We also believe that the definition of behavioral monitoring in this context should be broader, and new signals should be incorporated in the defense side to reason about the intent of remote entities in addition to the above traffic-based heuristics. 
% In section \ref{sec:in-application_sensors}, we explain how we can add systematic in-application sensors to improve visibility over the run-time activity.  
%In particular, in-application approaches such as \thesystem that provide fine-grained data about  spatio-temporal characteristics of web session can be of great benefit to complement existing defense methods. %We believe that services similar to the \thesystem project that can reason about the intent of remote agents in a more systematic and principled fashion
%is critical in protecting the trustworthiness of web applications and should be integrated into any web application that offer a critical service or handle important data. 
 %In the following, we provide more details on the threat model we use in this paper.  
 %\vspace{-.08in}
\item \textbf{Challenge-Based Defense Mechanisms.} As discussed in the introduction, solving a CAPTCHA challenge has been historically one of the main layers of defense against automated scanners and offensive bots. Suspicious agents are required to solve a CAPTCHA challenge by extracting the content of a distorted image, object, or audio file to access the desired service.
The underlying assumption is that human users can extract letters or identify objects more easily than scanners and non-human agents. 
Image-based CAPTCHA, reCAPTCHA \cite{von2004telling,von2008recaptcha}, reCAPTCHA v2~\cite{google-recaptcha} and hCAPTCHA~\cite{hcaptcha} are some of the main techniques used in practice.
%\noindent \textbf{Solving the Psychological Acceptability Challenge.}
%The original CAPTCHA-based systems such as reCAPTCHA \cite{von2004telling,von2008recaptcha}, 
%and Google's reCAPTCHA v1 and v2 utilized image, text, and audio recognition tasks.
However, these were effectively broken using recent advances in employing machine learning techniques \cite{alqahtani2020image,stark2015captcha,george2017generative,bursztein2014end,bock2017uncaptcha}. Additionally, a main defensive objective, in addition to detecting automated agents, is threat attribution and monitoring. To elaborate, gaining a holistic view of the threat landscape requires one to perform multi-class classification of automated agents into various categories, and to record and monitor each class's interactions. This provides
comprehensive insights into how remote agents interact and about their potential objectives, which is missing in conventional challenge-based defenses. 
\end{itemize}
\subsection{Machine Learning Models for Detection and Attribution}

 The recent trend towards using behavioral patterns for web scanner detection has been facilitated by the advances in machine learning and particularly deep learning techniques, e.g., \cite{suchacka2021efficient, tanaka2020bot}.  Furthermore, Google's reCAPTCHA v3 uses machine learning techniques to perform adaptive risk analysis in the background \cite{google_recaptcha_v3}. Similarly, the BeCAPTCHA \cite{acien2022becaptcha}, the Frs-sifs session identification technique of \cite{hamidzadeh2023frs}, user authentication methods of \cite{mondal2017study, niu2023exploring, wei2019deep} leverage convolutional and recurrent neural networks architectures for bot detection and authentication. 
In this work, we have utilized several machine learning architectures for real-time web scanner detection, offline attribution, and trend discovery:
\begin{itemize}[leftmargin=*]
    \item \textbf{Long-Short-Term-Memory (LSTM) Architectures.} Recurrent neural networks with long-short-term-memory  \cite{graves2005framewise} have been deployed successfully in various pattern recognition tasks such as handwriting recognition,  acoustic modeling, and protein secondary structure prediction, among others \cite{greff2016lstm}. The core concept of the LSTM structure revolves around a memory cell that can retain its status throughout time, accompanied by nonlinear gates that manage the transfer of information entering and exiting the cell. The internal memory makes LSTMs a suitable candidate for tasks with sequential inputs, such as bot detection based on behavioral biometrics data over a given time interval \cite{niu2023exploring}. We have used LSTMs in designing our real-time detection mechanisms in Section \ref{sec:class}. 
    \item \textbf{Hidden Markov Models (HMM).} An HMM represents a discrete-time, finite-state Markov chain, observed through a discrete-time memoryless invariant channel \cite{ephraim2002hidden}. That is, instead of observing the current state of the Markov chain at a given time in a (discrete output) HMM, only a function of the current state is observed.  
    HMMs have been widely used in the past several decades in various statistical modeling and hypothesis testing applications including speech recognition, natural language processing, robotics, and anomaly detection \cite{mor2021systematic}. A shortcoming of standard HMM techniques in statistical modeling is that they are only applicable to sequential scalar inputs, and are not amiable to modeling high dimensional input data. As a result, although HMM performance is comparable to more recent LSTM methods when only using uni-modal data such as mouse movement dynamics alone (Section \ref{sec:emp}), their performance is inferior when the input data is high dimensional, or has a large alphabet size (Section \ref{sec:th}) such as when multi-modal data is given. We use HMMs as a benchmark in our real-time bot detection simulations and as one of the components in our offline attribution and trend discovery mechanisms.
    \item \textbf{Spectral Clustering for Offline Attribution.}  Spectral clustering is a technique derived from graph theory, where instances are treated as vertices of a graph. It uses the eigenvalues of the Laplacian of the similarity matrix of the data to cluster it into dissimilar clusters \cite{von2007tutorial}. We have used spectral clustering techniques in Section \ref{sec:clust} to perform offline unsupervised learning on the collected multi-modal traces from the \thesystem forensics engine and to illustrate the possibility of detecting new emerging behavioral patterns of automated engines using these techniques. 
    \item \textbf{Agglomorative Clustering for Offline Attribution.}  Agglomerative clustering is a type of hierarchical clustering method that starts with each data point as its own cluster and then iteratively merges the closest pair of clusters into a single cluster \cite{murtagh2014ward}. The process is repeated until there's only a single cluster left, encompassing all data points, or until a termination criterion is met. A main advantage of this method is that it solely relies on pairwise distances of the nodes in the clustering process. This makes it applicable in scenarios where only pairwise distances are accessible such as the ones considered in Section \ref{sec:clust}.
\end{itemize}

%% file: 03-Methodology.tex
\begin{comment}
\section{Research Questions}
\label{sec:Methodology}
Our research aims to answer three fundamental questions
\noindent \textbf{RQ1:} 
How can defenders incorporate a \textbf{generalizable method for defending} against evasive web scanning attempts? When fingerprinting mechanisms might fail and Captcha challenges can be bypassed, what would be the alternative
without involving users in the loop or changing the underlying logic of the target web applications? 
\noindent \textbf{RQ2:} 
Could the method under consideration effectively operate in a \textbf{real-time manner}, meaning that it can function with minimal or negligible delay, ensuring timely and immediate results?
\noindent \textbf{RQ3:} 
What mechanisms or techniques does the model employ to identify and detect novel or previously unseen behaviors, as well as detect any significant changes in the data distribution, commonly referred to as \textbf{data drift}?
We aim to answer these questions by developing in-application instrumentation sensors and customized spatio-temporal artifacts used to monitor the run-time behavior of remote agents.
In the following, we discuss the implementation details and collected artifacts used to analyze the mentioned questions.
\end{comment}
\section{\thesystem: Multi-Modal Forensics Engine}
\label{sec:thesystem}
This section describes the design and implementation of \thesystem, and its data collection capabilities. In subsequent sections, an accompanying machine learning framework is developed to enable real-time detection, offline attribution, and trend discovery using the collected data. 
As mentioned in the introduction, more recent techniques~\cite{chong2019user,hamidzadeh2023frs,mondal2017study,acien2022becaptcha} have incorporated behavioral biometrics, specifically mouse movement dynamics, to differentiate
human traffic from web bots. While this technique offers more visibility into the behavior of connecting agents at the application level, the emergence of automated frameworks mentioned in section~\ref{sec:bots} has enabled the evasion of such defensive mechanisms by replicating human mouse movement dynamics 
and automatically triggering mouse events. In \thesystem, we add a new lens into the dynamic behavior of connecting agents by collecting three data points in a real-time fashion: (1) spatial, (2) temporal, and (3) activity type. Spatial and activity type data provides insights into how remote agents perceive different visual components of the web application. This manifests itself as triggering particular events in specific areas of the page.
Temporal data can provide a valuable modality on how frequently an event is triggered. Furthermore, it allows to extract the sequence of event activities triggered on each web session. A more complete discussion of the type of data collected by \thesystem is provided in Section \ref{sec:artifacts}.
%We also support \ak{num of events} event types that covers almost all the browsing events available for interaction with a web application. 
%The forensics engine also records 
%other metadata information, such as URI, Object IDs in the DOM of the page (e.g.,
%form, textfield, password, checkbox, submit).
 %The JavaScript library automatically activates 43 event sensors inside the target web application during the page loading process. For each event (e.g., a click on the page, DOM update due to clicking on an element), a callback event is generated. The recorded artifacts consist of three modalities: (1) spatial, (2) temporal, and (3) textual data. 
\subsection{Integration, Artifact Collection, and Data Transmission}
To enable data collection, \thesystem seamlessly integrates a suite of sensors into various features of the target web application, regardless of the application's underlying code or programming language. The client-side component of this engine is implemented as a JavaScript library extension of \cite{Leiva13-tweb}, which can be easily embedded into any web application through a simple script tag. We have open-sourced the prototype of both the in-application sensors and the server-side configurations to foster community engagement and further development.
\subsubsection{Event Listening and Callback Generation}
\begin{figure}[!t]
    \centering    \includegraphics[width=\linewidth]{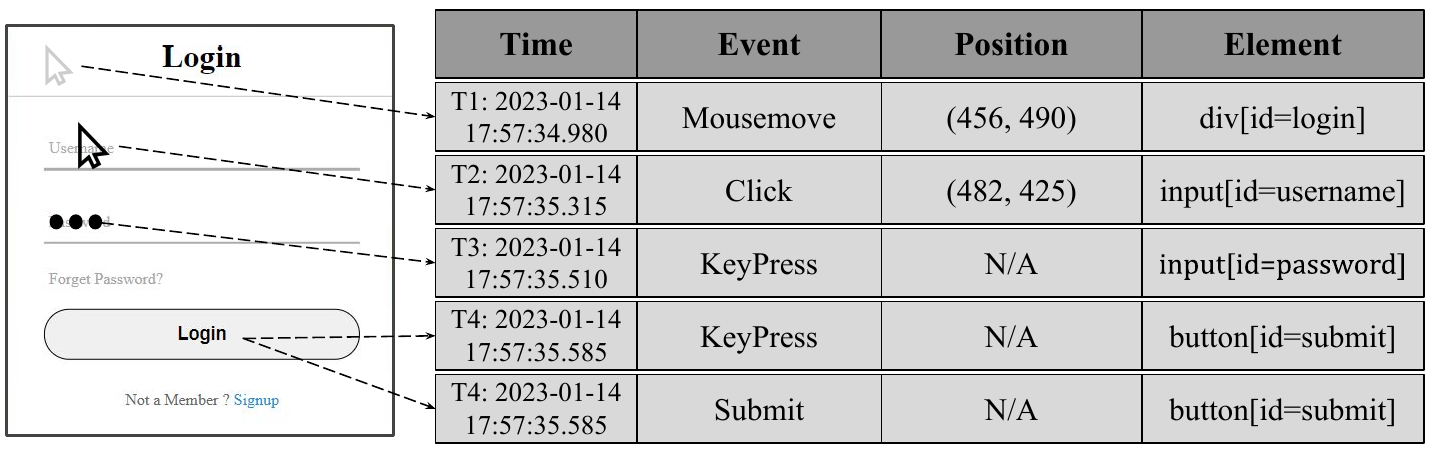}
    \caption{Sample Trace of the Recorded Events based on a Simplified Login Scenario}
    \label{tab:1}
\end{figure}

Once integrated, the JavaScript library automatically attaches event listeners to a broad array of features within the web application. Upon the activation of any such feature, a corresponding callback function is invoked. This callback captures and logs spatio-temporal data related to the event, detailing the timestamp and the location within the webpage where the event was triggered. Figure \ref{tab:1} provides an illustrative example of the granularity of the data logged for a session initiated by a human user.  As can be observed in Figure \ref{tab:1}, the user initiates a mouse movement at the time 
$T_1$, subsequently clicking on an input element at time $T_2$. The exact screen coordinates $(x,y)$ of the mouse pointer at both instances are captured as part of the event trace. Following this, the keypress event occurs at the time 
$T_3$, and eventually at time $T_4$, the user submits the form which triggers a keypress and a submit event. 
%  \begin{table*}
% \centering
% \caption{The collected spatio-temporal artifacts in a sample trace}
% \label{tab:1}
% \begin{tabular}{l|l}\hline 
% %\textbf{} & \multicolumn{3}{c}{\textbf{MacOS users}} & \multicolumn{3}{c}{\textbf{Windows users}}\\ \hline 
% \textbf{Timestamp} &\textbf{Trace Entry} \\\hline
% $T_{1}$: 2023-01-14, 17:57:34.980& action:\underline{mousemove} pos:(456,490) element:input  id:\textbf{question1} path:\textbf{./survey1}\\
% $T_{2}$: 2023-01-14 17:57:35.315& action:\underline{click} pos:(482,425) element:checkbox  id:\textbf{option1} path:\textbf{./survey1}\\
% $T_{3}:$ 2023-01-14 17:57:35.510& action:\underline{keypress} element:textarea  id:\textbf{answer2} path:\textbf{./survey1}\\
% $T_{4}:$ 2023-01-14 17:57:35.585 &action:\underline{submit} element: form id:\textbf{survey\_form} path:\textbf{./survey1} \\
% % & \textbf{Mean} & \textbf{Std. dev.} &\textbf{Obs} & \textbf{Mean} & \textbf{Std. dev.} 
% \hline
% \end{tabular} 
% \end{table*}

\subsubsection{Scope of Event Coverage}
\thesystem currently accommodates 44 event types (e.g., load, focus, mousemove, click, copy, scroll), which are widely supported across major browser platforms. Additionally, the engine captures relevant metadata, including the URL path, object IDs in the Document Object Model (DOM) (e.g., form, textfield, password, checkbox, submit), and the actions performed during the event. This provides a comprehensive snapshot of nearly all forensically pertinent events occurring during each web session. A complete list of the events collected by \thesystem is provided in \cite{evtrack_github}.
\subsubsection{Data Transmission Techniques and Efficiency}
\thesystem facilitates near real-time transmission of recorded event traces to a designated remote server. Traditional HTTP polling is one approach to achieve this real-time communication~\cite{http-polling}. However, HTTP polling has inherent limitations, such as increased network overhead attributed to repetitive HTTP header exchanges needed to maintain session state~\cite{http}. For a more complete discussion of this overhead, please refer to Section \ref{sec:overhead}. To address these shortcomings, we employ a WebSocket-based communication channel as an alternative.
WebSocket transmissions are highly efficient, as evidenced by prior empirical studies conducted in controlled settings~\cite{murley:2021:real-time, http}. Specifically, a WebSocket transmission necessitates, on average, 46 bytes of network bandwidth. This includes an 8-byte WebSocket header and an average 38-byte variable-sized binary payload.

\paragraph{Communication Overhead}
Our empirical evaluation, elaborated in Section \ref{sec:emp}, reveals an average event listener activation time approximating 10 milliseconds. This leads to an average communication overhead for \thesystem that is less than 10 KB per second. It should be noted that this overhead is contingent on multiple factors, such as the structure of the website and the number of activated event listeners.
\paragraph{Practical Implications of the Communication Overhead}
It's worth noting that the frequency of event listener activation --- and consequently the communication overhead --- is subject to the specific use-case scenario.
In typical applications, however, event listeners are unlikely to be triggered more frequently than once per millisecond. As a result, we estimate a practical upper limit for the communication overhead {at less than ten KB per second.}

\subsection{Data Collection and Empirical Evaluation}
\label{sec:artifacts}

To illustrate the efficacy and practical applicability of \thesystem, we implemented a real-world testbed: a survey website instrumented with \thesystem's analytics engine. This survey was part of an evaluation of five courses taught in the institution and provided us with authentic human interaction data. The human data spans a period starting from January 8, 2023, at 11:46:41 EST to February 27, 2023, at 19:57:23 EST. Additionally, we have collected data from automated scanners' interactions with the page. Beyond human-generated data, we also collected a dataset comprising interactions from automated agents. Specifically, these agents belonged to the various classes introduced in Section \ref{sec:bots}.

Each user session on the website generated an interaction trace through \thesystem. These traces encapsulate critical event attributes including but not limited to the nature of the event (e.g., click, scroll), the associated timestamp, and the screen coordinates where the event was triggered (see Figure~\ref{tab:1} for an example). Specifically, every trace entry includes an event key, which serves as a constant string identifier for the action, and one or more event values that provide additional context or parameters for the event, coupled with precise timestamps.

%\noindent \textbf{Crawler Bot.} For the bot interaction artifacts, we developed an automated web scanner running as full-fledged browser using Selenium \cite{Selenium}. The scanner crawls a target webpage, finds the available link tags on the page, visits each link, and comes back to the main page. In order to make the scanner's behaviour closer to humans, the scanner mouse moves on a line with artificial random delays of 10ms-30ms between each event to reach the target link. The target website for this task is the security course projects pages which includes links to outside resources and also other projects. We deployed 500 scanners and collected all of the events. 
\noindent \textbf{Data Protection and Permissions.} Before collecting human interactions, we submitted an application to the Institutional Review Board (IRB) and received a waiver to run the experiment. The experiment was launched anonymously. We did not collect any potentially identifiable information such as username, email address, or IP address in the course of the experiment. We acknowledge there might be some minimal risks to mapping enrolled students in those classes with those who completed the surveys, but we have not and do not intend to collect any metadata in any way based on the defined IRB protocol.
Data protection procedures have been implemented
at all the layers. The collected data was aggregated and encrypted via a public key on a local machine. The machine is only accessible to two members of the team who successfully passed special IRB training courses.

\noindent \textbf{Human Interaction Data.} The survey itself is composed of 40 heterogeneous questions, formatted as multiple-choice checkboxes, dropdown menus, and free-response text fields. To ensure that only real users can access the survey page and interact with it, we set up a password-enabled page that requested pre-shared credentials before redirecting to the actual page. To this end, one specific pair of username and password was shared among all the participants to ensure only real users could interact with the page while minimizing the amount of data about participants. The average duration of interaction per user was approximately 35 minutes. Across the data collection period, we successfully recorded 46 distinct interaction traces, comprising a rich dataset of 825,701 individual event artifacts.

\noindent \textbf{ZAP \cite{owasp_benchmark_2017} Interaction Data.} In order to collect interaction data from real vulnerability scanners, we utilized OWASP ZAP.   We deployed 104 instances of the scanner and collected the artifacts. In the collected dataset, each instance contains an average of more than 500 artifacts. Leading to a total of 58,014 artifacts.  
 
\noindent \textbf{HLISA Interaction Data.} We utilized HLISA \cite{gossen2021hlisa} to emulate a crawler that is close to human interactions. We deployed 500 scanners and collected all of the events. In the collected dataset, each instance contains an average of more than 900 artifacts. Leading to a total of 472,665 artifacts.

\noindent \textbf{Gremlins Interaction Data \cite{richard_2020}}
To collect the artifacts for another scanner that emulates human behavior, we utilized \textit{gremlins.js}~\cite{gremlins_github}. Prior work~\cite{trickel2023toss, azad2019less} have used the Gremlins framework to automatically generate realistic user activities in interaction with web applications as well. We deployed 500 instances of the gremlins separately and collected the artifacts. In the collected dataset, each instance contains an average of more than 400 artifacts. Leading to a total of 219,524 artifacts. 

 \noindent \textbf{Random Bots Interaction Data Using Puppeteer.} We developed a set of web scanners using Puppeteer \cite{puppeteer} which open a target URL, select random positions on the screen based on a Gaussian distribution, and move the mouse to that location with predefined movement patterns. The movement patterns include naive movement and artificially delayed movement.
 The bot repeats the process for ten seconds and exits. We deployed 100 instances of the naive moving bot and 500 instances of the artificially delayed bot and collected the artifacts from their interactions. In the collected dataset, each instance contains an average of more than 800 and 400 artifacts for the naive and artificially moving bots, respectively. Leading to a total of 85,940 and 219,524 artifacts, respectively. 
 
 A summary of the statistics for each of the collected datasets is provided in Table \ref{tab:ds}.

\begin{table}[h!]
\caption{Summary of Collected Interactions from Various Agents}
\label{tab:ds}
\resizebox{\columnwidth}{!}{%
\begin{tabular}{llrr}
\toprule
\textbf{Agent Type} & \textbf{Name} & \textbf{Number of Traces} & \textbf{Total Number of Artifacts} \\
\midrule
Human & Survey & 46 & 825,701 \\ 
\midrule
UI Testing Bot & Gremlins.js \cite{richard_2020} & 500 & 219,524 \\
\midrule
Crawler Bot & HLISA \cite{gossen2021hlisa} & 500 & 472,665 \\
\midrule
Scanner Bot & OWASP ZAP \cite{owasp_benchmark_2017} & 104 & 58,014 \\
\midrule
Random Bot & Naive Movement & 100 & 85,940 \\
\midrule
Random Bot & Artificially Delayed & 500 & 219,524 \\
\bottomrule
\end{tabular}%
}
\end{table}

\section{Machine Learning Framework for Online Detection and Attribution}
\label{sec:ml}

In the preceding sections, we described our implementation of \thesystem which enables real-time data collection by monitoring user activities in a web application and extracting a set of relevant features. In this section, we focus on developing a complementary analytical framework that supports both offline clustering and online classification of web application visitors using these collected features.
The objective is to classify visitors into benign human users and classes of potentially malicious automated agents that could be responsible for scraping, credential stuffing, DDoS,  
and content automation among others.
Generally, we wish to perform:
\begin{enumerate}[left=0pt]
\item \textbf{Offline Attribution and Trend Discovery:} Observe the website visitors for an extended period of time (e.g., several days) and perform unsupervised learning on the collected data to identify different visitor classes.

\item \textbf{Online Detection and Classification:} Observe new visitors 
for a short time period (e.g., hundreds of milliseconds) and perform real-time multiclass hypothesis testing to detect users belonging to each of the classes identified in the clustering step.
\end{enumerate} 
To accomplish this, in Section \ref{sec:clust}, we outline an offline clustering approach based on  HMMs, and utilizing agglomerative clustering and spectral clustering techniques \cite{bishop2006pattern}.   This allows the website administrator to perform unsupervised clustering and identify unique classes of benign users and malicious agents visiting the website over a given time frame. In Section \ref{sec:class}, we detail methods for multi-class hypothesis testing and supervised learning techniques to classify new users into the classes identified in Section \ref{sec:clust} through unsupervised learning techniques. 

\subsection{Data Preprocessing}
\label{sec:pre}
As a first step, we preprocess the data to make it amiable to application in supervised and unsupervised learning scenarios considered in subsequent sections. To this end, we wish to map each trace element into a vector in a finite-dimensional Euclidean space. We follow the following preprocessing steps:

\begin{table}
\centering
\caption{Processed Spatio-temporal Artifacts}
\label{tab:2}
\begin{tabular}{l|l}
\hline
\textbf{Key} & \textbf{Value} \\
\hline
\texttt{'x'} & \texttt{[456, 482, 482, 482, 482]} \\
\texttt{'y'} & \texttt{[490, 425, 425, 425, 425]} \\
\texttt{'t'} & \texttt{['2023-01-14,17:57:34.980',} \\
           & \texttt{'2023-01-14 17:57:35.315',} \\
           & \texttt{'2023-01-14 17:57:35.510',} \\
           & \texttt{'2023-01-14 17:57:35.585',} \\
           & \texttt{'2023-01-14 17:57:35.585']} \\ \hline
\texttt{'event'} & \texttt{[2, 14, 13, 13, 19]} \\
\texttt{'vel'} & \texttt{[0, 3798.0, 0, 0, 0]} \\
\texttt{'dir'} & \texttt{[0, 3, 0, 0, 0]} \\
\texttt{'dt'} & \texttt{[Timedelta('0 days 00:00:00.000'),} \\
               & \texttt{Timedelta('0 days 00:00:00.335'),} \\
               & \texttt{Timedelta('0 days 00:00:00.195'),} \\
               & \texttt{Timedelta('0 days 00:00:00.075'),} \\
               & \texttt{Timedelta('0 days 00:00:00.000')]} \\
\hline
\end{tabular}
\end{table}

\begin{itemize}[leftmargin=*]
    \item \textbf{Event Enumeration:} Each event monitored by \thesystem is assigned an integer index.
    
    \item \textbf{Velocity and Inter-arrival Time Calculation:} We calculate the velocity \( v \) of each mouse movement as \( (v_x,v_y) = (\Delta x,\Delta y) / \Delta t \), where \( \Delta x \)  and \(\Delta y\) are the change in position in the x and y directions, respectively, and \( \Delta t \) is the change in time. We also compute the inter-arrival times \( \tau \) between events. If Hidden Markov Models (HMMs) are used, the velocity values are further discretized through quantization, whereas LSTMs allow for continuous-valued inputs and do not require the quantization step. In the simulations in Section \ref{sec:emp}, we have employed two bins for the quantization of each of the x-direction and y-direction velocities. The choice of quantization thresholds is determined via hyperparameter optimization. Lastly, the discretized two-dimensional vector is mapped to a one-dimensional vector (denoted by $\mathtt{dir}$) through an isomorphic transformation.  
    
    \item \textbf{Handling Missing Location Values:} As can be observed in our collected dataset, for some events, the mouse location might not be logged. In these cases, we backfill the missing data with the most recent available mouse location.
    
    \item \textbf{Accounting for Long Inter-arrival Times:} The Long Short-Term Memory (LSTM) model described in Section \ref{sec:class} can accept continuous-valued, multidimensional input vectors. However, HMMs are more suited for discrete input. Thus, inter-arrival times are discretized using non-uniform quantization. The quantization bins are constructed such that each contains an equal number of trace elements. Each data entry is then replicated a number of times corresponding to the quantization index of its inter-arrival time. The total number of quantization bins can be chosen by hyperparameter optimization.
    
    \item \textbf{Scalarization:} While LSTM models are capable of processing multidimensional data, HMMs are more amenable to scalar inputs. Accordingly, when using an HMM, the multidimensional, discrete-valued vector obtained from previous preprocessing steps is mapped to a scalar value through an isomorphic transformation. For better comprehension of these preprocessing steps, Table \ref{tab:2} illustrates the resultant output when the trace elements from Figure \ref{tab:1} are processed. The upper segment of Table \ref{tab:2} shows the information of the collected sample artifacts and the bottom section includes different calculated features that are being used directly in the analysis.
\end{itemize}

\subsection{Ground Truth Assumptions}
\label{sec:assu}
Our foundational assumptions in modeling the classification and clustering tasks are described in this section. Let $\mathcal{U}=\{u_1,u_2,\cdots,u_n\}$ denote the set of all web application visitors, where $n\in \mathbb{N}$. For a given $i\in [n]$, we write  \(\mathbf{X}^{(i)} = \{X^{(i)}_1, X^{(i)}_2, \ldots\}\) to represent the discrete-valued random process obtained from interactions of user $u_i$ following the preprocessing methodology described in Section \ref{sec:pre}.
\\\textbf{Assumption 1 (Ergodicity):} For any $i\in [n]$, we assume that the random process \(\mathbf{X}^{(i)}\) is an (almost stationary) ergodic process. The probability measure associated with \(\mathbf{X}^{(i)}\) is denoted by \(\mu^{(i)}\), and the joint distribution of \(X^{(i)}_1, X^{(i)}_2, \ldots, X^{(i)}_t\) is denoted by \(\mu^{(i)}_t\), where \(t \in \mathbb{N}\).
\\\textbf{Assumption 2 (Clusterability):} For any three arbitrarily selected users with associated probability measures $\mu,\mu'$ and $\mu''$, where the first two users belong to  the same class, and the third user belongs to a different class, we assume: 
\begin{align}
\label{eq:2}
    P\left(D_J(\mu_{t} \parallel \mu'_t) > D_J(\mu_{t} \parallel \mu''_t)\right)\to 0, \text{ as } t\to \infty,
\end{align}
where  \(D_J(\cdot \parallel \cdot)\) is the Jeffreys divergence (also called the symmetrized Kullback-Leibler divergence). The Jeffreys divergence between two measures $\mu$ and $\eta$ is defined as $\mathbb{E}_{\mu}(\log{\frac{\mu}{\eta}})+\mathbb{E}_{\eta}(\log{\frac{\eta}{\mu}})$.

Equation \eqref{eq:2} essentially states that users belonging to the same class are closer to each other in the Jeffreys divergence sense compared to those in other classes. As a result, the clustering task can be accomplished by finding estimates of the underlying probability measure for each user, and grouping users together --- for example, using spectral or agglomerative clustering techniques --- based on the pairwise Jeffreys divergences of the estimated distributions.

\textbf{Assumption 3 (Class Representatives):} Consider the set of ground-truth labels $\mathcal{C} $ $\triangleq \{c_1, c_2, \ldots, c_{k}\}$, where \(c_1\) corresponds to human users and \(c_i, i\neq 1\), correspond to various types of automated agents. We assume the existence of class representative measures \(\mu_{1}, \mu_{2}, \ldots, \mu_{k}\), such that for any given user, the following holds:
\begin{align}
\label{eq:1}
    P\left(D_J(\mu_{i,t} \parallel \mu_t) > D_J(\mu_{j,t} \parallel \mu_t) \mid \mathcal{E}_i\right)\to 0, \text{ as } t\to \infty,
\end{align}
for all \(i \neq j\), where \(\mathcal{E}_i\) is the event that the user belongs to class \(c_i\), the probability measures  \(\mu_{i,t}, \mu_t, \mu_{j,t}\) denote the \(t\)-letter probabilities corresponding to the class representative of \(c_i\), the user, and class representative of \(c_j\), respectively.

% \bo{Maybe only use the following for this section. and remove the extra math stuff.}

Equation \eqref{eq:1} essentially states that, as the length of observed user traces increases, the distribution of the observed sequence becomes closer in divergence to the distribution of its corresponding class representative, compared to any other class representative. With this assumption, the classification task can be broken into two parts; first, we derive an estimate \(\hat{\mu}_1, \hat{\mu}_2,\cdots, \hat{\mu}_{k}\) of the class representative distributions using a labeled training set. This can be completed using an HMM or LSTM as described in subsequent sections. Next, for any new user visiting the web application, an estimate of its underlying distribution $\hat{\mu}$ is derived and the user is classified by finding its closest class representative in Jeffreys divergence sense.

It should be noted that Assumption 3 is not a consequence of Assumption 2 and vice versa as Jeffreys divergence does not satisfy the triangle inequality. 

\subsection{Offline Attribution and Trend Detection}
\label{sec:clust}
%In the offline attribution step, the web application traffic is monitored for an extended period of time, and unlabeled traces of user activities are logged. The objective is to perform unsupervised learning via clustering techniques to find the different classes of human users and automated agents which have visited the website. This labeled data is used as training data in the classification step to perform supervised learning. 
As discussed in Section \ref{sec:assu}, in order to perform clustering, for each observed trace of user behaviors, we train an HMM to estimate the underlying probability measure. Then, we use a clustering technique to group traces together according to their pairwise Jeffrys divergences. 
HMMs are widely used in modeling generative sequences, characterized by an underlying (hidden) Markovian process. This hidden process generates an observable sequence of outcomes, such that the observed outcomes are conditionally independent of one another, given the hidden process. 
%Generally, HMMs have been used to study speech recognition, natural language processing, robotics, and computer vision, among various other applications
%\cite{mor2021systematic}. Particularly related to this work, HMMs have been used in  classification and prediction tasks in evaluating human intent by observing physical mobility \cite{kelley2008understanding}, navigation tasks \cite{zhu1991hidden}, trajectory learning \cite{vakanski2012trajectory}, and human action learning \cite{yang1997human}.
Formally an HMM is defined as follows.

\begin{definition}[\textbf{Hidden Markov Processes}]
Consider a sequence of random variables \(\{X_i\}_{i\in \mathbb{N}}\) with a finite alphabet \(\mathcal{X}\). This sequence is a Hidden Markov Process if there exists a finite state space \(\mathcal{S}\), a distribution \(P_{S}\), and conditional distributions \(P_{S'|S}: \mathcal{S}\times \mathcal{S} \to [0,1]\) and \(P_{X|S}:\mathcal{X}\times\mathcal{S}\to [0,1]\), such that:
\begin{align}
\label{eq:3}
    P_{X^n}(e^n) &= \sum_{s^n\in \mathcal{S}^n} P_S(s_1)\prod_{i=2}^n P_{S'|S}(s_i|s_{i-1})\prod_{i=1}^n P_{X|S}(x_i|s_i),
\end{align}
for all \(n\in \mathbb{N}\) and \(x^n\in \mathcal{X}^n\). The process is parameterized by the tuple \((\mathcal{X},\mathcal{S},P_{S'|S}, P_{X|S})\). The sequence of random variables \(\{S_i\}_{i\in \mathbb{N}}\) are called the (hidden) states, and \(\{X_i\}_{i\in \mathbb{N}}\) are called the emissions.
\end{definition}
\noindent\textbf{Example.} Assume that we use a forensics engine to monitor the activities of a website visitor. Suppose there are three possible events: 1) left mouse movement, 2) right mouse movement, 3) mouse click. These observations can be modeled as an HMM by mapping them to a finite set of integers \(\mathcal{X} = \{1, 2, 3\}\). For example, the observation sequence `left, left, left, right, click, right' can be mapped to \(X_1=1,X_2=1,X_3=1,X_4=2,X_5=3,X_6=2\). The hidden states \(S_i, i\in \mathbb{N}\) might represent underlying user intentions or behaviors. The cardinality of the hidden state alphabet and the stochastic relation between emissions and hidden states can be estimated using empirical observations of past emission sequences  (e.g., \cite{yang2017statistical,ephraim2002hidden}).

To perform clustering, for each user $u_i$, 
with observed trace $\mathbf{X}^{(i)}$, we use the widely used Baum-Welch algorithm \cite{baum1972inequality} to find estimates $\widehat{P}^{(i)}_{X|S}$, $\widehat{P^{(i)}}_{S_2|S_1}$, and $\widehat{P}^{(i)}_S$. The number of hidden states $s$ is determined by hyperparameter optimization (e.g., \cite{ephraim2002hidden}). 
Then, for a hyperparameter $t$, we find the Jeffrys divergence between $t$-letter probability measures $D_J(P^{(i)}_{{X^t}}||P^{(j)}_{{X^t}}), i,j\in [n]$, where $P_{X^t}$ is derived using Equation \eqref{eq:3}. The pairwise divergences are fed to a clustering method to group the users together. The clustering algorithm is provided in Alg. \ref{alg:clustering}.

\begin{algorithm}[t]
\caption{Clustering Algorithm}
\label{alg:clustering}
\begin{algorithmic}[1]
    \STATE {\bfseries Input:} user set $\mathcal{U}$; traces $\mathbf{X}^{(i)}$;  Hyperparameters $t,s,k$.
    \STATE {\bfseries Output:} $\mathcal{U}_1,\mathcal{U}_2,\cdots, \mathcal{U}_k$ clusters of users.
    \STATE \textbf{Initialization:}
    \STATE $\text{DivMat}$ \textcolor{gray}{ \COMMENT{\#$n\times n$ matrix to store pairwise divergences}}
    \STATE \textbf{Compute HMM Initial and Transition Probabilities:}
    \FOR{each $u_i \in \mathcal{U}$}
        \STATE $\widehat{P}^{(i)}_{S}, \widehat{P}^{(i)}_{S_2|S_1}, \widehat{P}^{(i)}_{X|S}\leftarrow BaumWelch(\mathbf{X}^{(i)},s)$ 
    \ENDFOR
    \STATE \textbf{Compute Pairwise Divergence:}
    \FOR{$i=1,2,...,n$}
        \FOR{$j=i+1,i+2,...,n$}
            \STATE $P^{(i)} \leftarrow P^{(i)}_{{X^t}}$ \textcolor{gray}{ \COMMENT{\# Compute $P^{(j)}_{{X^t}}$ via Eq.\eqref{eq:3}}}
            \STATE $P^{(j)} \leftarrow P^{(j)}_{{X^t}}$  \textcolor{gray}{\COMMENT{\# Compute  $P^{(j)}_{{X^t}}$ via Eq.\eqref{eq:3}}}
            \STATE $d \leftarrow D_J(P^{(i)} \parallel P^{(j)})$ \textcolor{gray}{\COMMENT{\# Compute Jeffreys Divergence}}
            \STATE  $\text{DivMat}[i][j] \leftarrow d$
            \STATE  $\text{DivMat}[j][i] \leftarrow d$
        \ENDFOR
    \ENDFOR
    \STATE \textbf{Clustering:}
    \STATE Apply spectral or agglomerative clustering on $\text{DivMat}$ to form clusters $\mathcal{U}_1,\mathcal{U}_2,\cdots, \mathcal{U}_k$.
    \RETURN $\mathcal{U}_1,\mathcal{U}_2,\cdots, \mathcal{U}_k$  
\end{algorithmic}
\end{algorithm}
%In the sequel, we investigate the outputs generated by our forensics engine, assuming that they are produced based on an HMM. Using prior observations, we empirically estimate the HMM's parameters, specifically obtaining estimates \(\widehat{P}_{S'|S}\) and \(\widehat{P}_{X|S}\) for the transition probabilities \(P_{S'|S}\) and emission probabilities \(P_{X|S}\), respectively. The granularity of our observations affects the alphabet size \(|\mathcal{X}|\). On one hand, more a granular observation scheme allows for greater information collection, potentially enhancing the accuracy of the clustering and classification tasks. On the other hand, it necessitates a larger number of training samples to obtain accurate estimates of \(P_{S'|S}\) and \(P_{X|S}\). This is made precise in the following proposition. 

\subsection{Online Detection and Classification}
\label{sec:class}

 In the  classification task, we wish to observe new visitors for a short time period (e.g., hundreds of milliseconds) and perform real-time
multiclass hypothesis testing to detect users belonging to
each of the classes identified in the clustering step. To perform the classification task, we consider several hypothesis tests, using Long Short Term Memory (LSTM) architectures as the main model, and Hidden Markov Models (HMM)s as a baseline for comparison. 

\begin{itemize}[leftmargin=*]

\item{\textbf{LSTM Classifier.}} The LSTM architecture is capable of taking continuous-valued vector inputs and does not require the (information-lossy) scalarization and discretization steps applied to produce the HMM input in the previous section. 

\textbf{Definitions:} For a given user \(u_i\), let the session start and end times be denoted by \(t^{(i)}_{\text{init}}\) and \(t^{(i)}_{\text{end}}\), respectively. Define:
\begin{align*}
&\mathbf{T}^{(i)} \triangleq \{ T_i \mid  T_i \in [t^{(i)}_{\text{init}}, t^{(i)}_{\text{end}}] \},
\quad 
\mathbf{C}^{(i)} \triangleq \{ (x_t, y_t) \mid t \in \mathbf{T}^{(i)} \},
\\&
\mathbf{E}^{(i)} \triangleq \{ e_t \mid t \in \mathbf{T}^{(i)} \}.
\end{align*}
\textbf{Training Phase:} Let \(\mathcal{U}_j, j \in [k]\) denote the set of user clusters identified in the clustering phase. Define:
\begin{align*}
&\mathbf{T}_j \triangleq \{\mathbf{T}^{(i)} \mid  u_i \in \mathcal{U}_j\},
\quad
\mathbf{C}_j \triangleq \{(\mathbf{C}_x^{(i)}, \mathbf{C}_y^{(i)}) \mid u_i \in \mathcal{U}_j\},
\\&
\mathbf{E}_j \triangleq \{\mathbf{E}^{(i)} \mid u_i \in \mathcal{U}_j\}.
\end{align*}
The labeled tuples \((\mathbf{T}_j, \mathbf{C}_j, \mathbf{E}_j)\) are sequentially fed to the LSTM in the training phase.
\\\textbf{Evaluation Phase:} Consider a new user \(u\) visits the website at time \(t_{\text{init}}\). At a given time \(t\), the LSTM takes as input:
\begin{align*}
&
\mathbf{T} \triangleq \{ T_i \mid T_i \in [t_{\text{init}}, t] \},\quad 
\mathbf{C} \triangleq \{ (x_t, y_t) \mid t \in \mathbf{T} \},
\\&
\mathbf{E} \triangleq \{ e_t \mid t \in \mathbf{T} \}.
\end{align*}
Let the output label be denoted by \(\hat{c}\). The LSTM continues classification until the output \(\hat{c}\) is repeated \(q\) number of times, where \(q \in \mathbb{N}\) is a hyperparameter, which is chosen depending on the desired accuracy vs detection time trade-off and is discussed in more detail in subsequent sections.

The training and evaluation steps are shown in Alg. \ref{alg:lstm_classifier}.

\begin{algorithm}[t]
\caption{LSTM Classifier Algorithm}
\label{alg:lstm_classifier}
\begin{algorithmic}[1]
    \STATE {\bfseries Input:} Clustered user sets $\mathcal{U}_1,\mathcal{U}_2,\cdots, \mathcal{U}_k$; user traces $\mathbf{T}_j$, $\mathbf{C}_j$, $\mathbf{E}_j$; Hyperparameter $q \in \mathbb{N}$.
    \STATE {\bfseries Output:} Classification label $\hat{c}$ for new user.
    \STATE \textbf{Training Phase:}
    \FOR{each $\mathcal{U}_j, j \in [k]$}
        \FOR{each tuple $(\mathbf{T}^{(i)}, \mathbf{C}^{(i)}, \mathbf{E}^{(i)})$ where $u_i \in \mathcal{U}_j$}
            \STATE $\mathtt{LSTM.fit} (\mathbf{T}^{(i)}, \mathbf{C}^{(i)}, \mathbf{E}^{(i)})$ 
        \ENDFOR
    \ENDFOR

    \STATE \textbf{Evaluation Phase:}
    \STATE Initialize $\text{count} = 0$ and $c_{temp}=-1$.
    \WHILE{True}
    \IF {$\mathbf{T}\neq \{ T_i \mid T_i \in [t_{\text{init}}, t] \}$}
        \STATE $(\mathbf{T},\mathbf{C},\mathbf{E}).concatenate(T_t,C_{t},E_t)$
        \STATE $\hat{c}\leftarrow
        \mathtt{LSTM.predict}(\mathbf{T},\mathbf{C},\mathbf{E})$
        \ENDIF
        \STATE $count\leftarrow \mathds{1}(c_{temp} = \hat{c})(count +1)$
        \STATE $c_{temp}\leftarrow \hat{c}$
        \IF{$count \geq q$}
            \RETURN  $\hat{c}$.
        \ENDIF
    \ENDWHILE
\end{algorithmic}
\end{algorithm}
\item \textbf{HMM Classifier:} In addition to the LSTM classifier, we also train an HMM classifier, to serve as a  comparative benchmark for evaluating the LSTM's performance.
\\\textbf{Training Phase:} Define the set of labeled tuples, $\mathbf{X}_j$, as
\begin{align*}
\mathbf{X}_j \triangleq \{\mathbf{X}^{(i)} \mid u_i \in \mathcal{U}_j\}.
\end{align*}
 These labeled tuples are fed into the Welch-Baum algorithm to train the HMM models, denoted as $HMM_j$, one corresponding to each label $j$.
\\\textbf{Evaluation Phase:} Suppose a new user $u$ visits the website at an initial time $t_{\text{init}}$. At any subsequent time $t$, the HMM classifier takes as input the observation sequence $\mathbf{X} = \{X_i \mid t_i \in [t_{\text{init}}, t]\}$. The classifier computes the log-likelihood of this observed sequence under each of the trained $HMM_j$ models. Let these log-likelihoods be denoted by
$        \ell_{t,j} \triangleq \log{P_{t,j}}$,
    where $P_{t,j}$ is the likelihood of the observed sequence given the $HMM_j$ model. The classifier then calculates the difference metric, $\Delta_t$, as
$        \Delta_t \triangleq \max_{j} \min_{i} (\ell_{t,j} - \ell_{t,i})$.
    If $\Delta_t > \gamma$, where $\gamma$ is a predefined threshold hyperparameter, the classifier outputs the label corresponding to $
        \arg\max_{j} \min_{i} (\ell_{t,j} - \ell_{t,i})$.    This threshold $\gamma$ may be fine-tuned to optimize the trade-off between accuracy and time-to-detection. Note that this multiclass hypothesis test is known to be asymptotically optimal when the input data is generated according to an HMM model~\cite{draglia1999multihypothesis}. The operations of the classifier are summarized in Alg. \ref{alg:hmm_classifier}. 
\end{itemize}

\begin{algorithm}[h]
\caption{HMM Classifier Algorithm}
\label{alg:hmm_classifier}
\begin{algorithmic}[1]
    \STATE {\bfseries Input:} Clustered user sets $\mathcal{U}_1, \mathcal{U}_2, \ldots, \mathcal{U}_k$; Hyperparameter $\gamma \in \mathbb{R}$.
    \STATE {\bfseries Output:} Classification label $\hat{c}$ for new user $u$.
    
    \STATE \textbf{Training Phase:}
    \FOR{each $\mathcal{U}_j, j \in [k]$}
        \FOR{each $\mathbf{X}^{(i)}$ where $u_i \in \mathcal{U}_j$}
            \STATE $HMM_j \leftarrow \mathtt{WelchBaum.fit}(\mathbf{X}^{(i)})$
        \ENDFOR
    \ENDFOR

    \STATE \textbf{Evaluation Phase:}
    \STATE Initialize $t_{\text{init}}$ when user $u$ visits the website
    \STATE Initialize $\mathbf{X} = \emptyset$
    \WHILE{True}
        \IF{$\mathbf{X}\neq \{X_i|t_i\in [t_{init},t]\}$}
            \STATE $\mathbf{X}.\mathtt{concatenate}(X_t)$
            \FOR{each $j \in [k]$}
                \STATE $\ell_{t,j} \leftarrow \log \left( \mathtt{HMM}_j.\mathtt{predict\_likelihood}(\mathbf{X}) \right)$
            \ENDFOR
            \STATE $\Delta_t \leftarrow \max_{j} \min_{i} (\ell_{t,j} - \ell_{t,i})$
            \IF{$\Delta_t > \gamma$}
                \STATE $\hat{c} \leftarrow \arg \max_{j} \min_{i} (\ell_{t,j} - \ell_{t,i})$
                \RETURN $\hat{c}$
            \ENDIF
        \ENDIF
    \ENDWHILE
\end{algorithmic}
\end{algorithm}

%% file: 04-Analysis.tex
\section{Empirical Analysis}
\label{sec:emp}
To verify the efficacy of \thesystem, in this section, we provide various simulations of the clustering and classification algorithms of Section \ref{sec:ml} applied to the real-world data described in Section \ref{sec:artifacts}.
\subsection{Clustering}
\begin{figure*}[!t]
  \centering
  \begin{minipage}{.5\textwidth}
    \centering    \includegraphics[width=.8\linewidth]{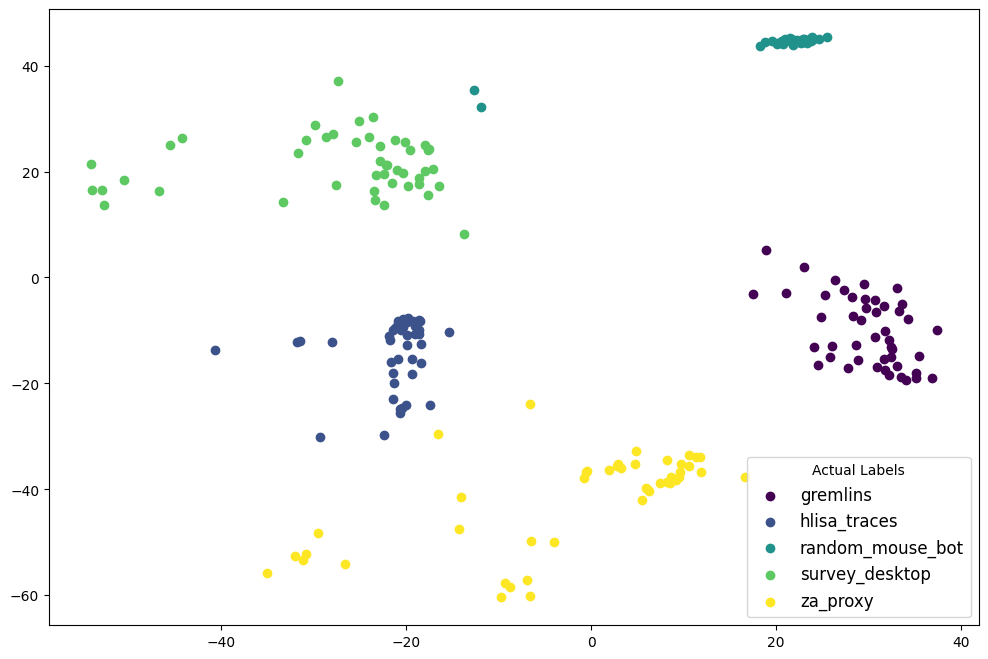}
  \end{minipage}%
  \hfill
  \begin{minipage}{.5\textwidth}
    \centering
    \includegraphics[width=.8\linewidth]{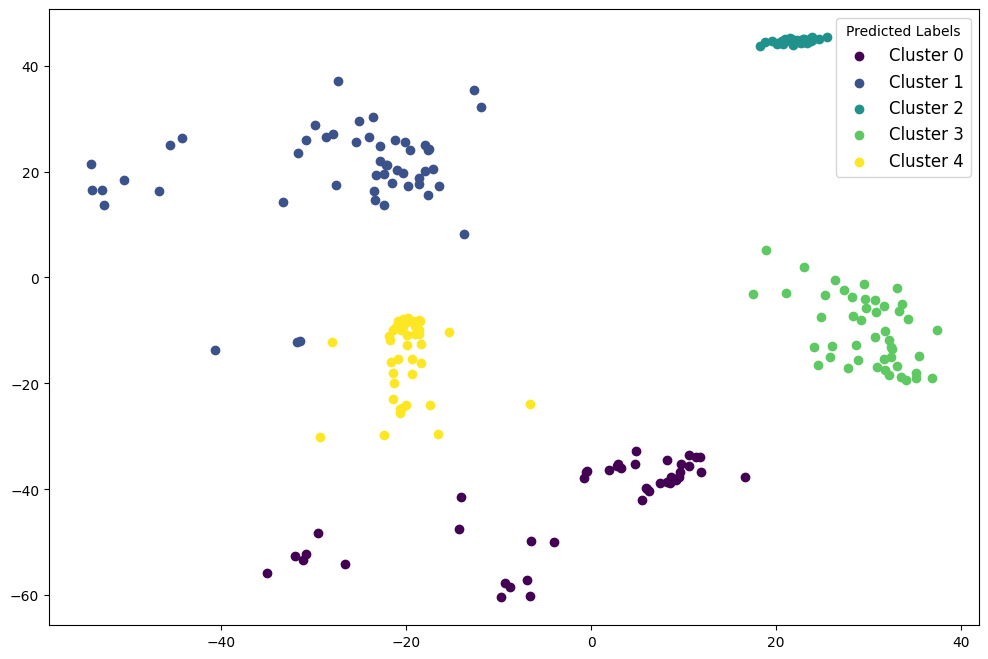}
  \end{minipage}
  \caption{Left: the actual labels of the input traces. Right: the output estimated labels using Algorithm \ref{alg:clustering}.}
      \label{fig:cl1}
\end{figure*}

To perform clustering, as described in Algorithm \ref{alg:clustering}, we generate one HMM model for each trace of unlabeled data. We take four hidden states for the HMM model, and train the model on at most 10,000 trace elements, discarding the rest. We train a total of $230$ HMM models, on 46 traces chosen from each of the categories `Human', `HLISA', `Gremlins', `ZAP', and `Random Bot'. We have chosen 15 quantization bins to quantize the time intervals, and 3 quantization bins to quantize the velocities in each direction as described in Section \ref{sec:pre}. We have used Spectral Clustering with a Gaussian Kernel and parameter $\sigma=9$.
Figure \ref{fig:cl1} shows the original and estimated labels. The clustering scores are provided in Table \ref{tab:clustering_metrics}. 

\begin{figure*}[!t]
  \centering
  \begin{minipage}{.5\textwidth}
    \centering    \includegraphics[width=.8\linewidth]{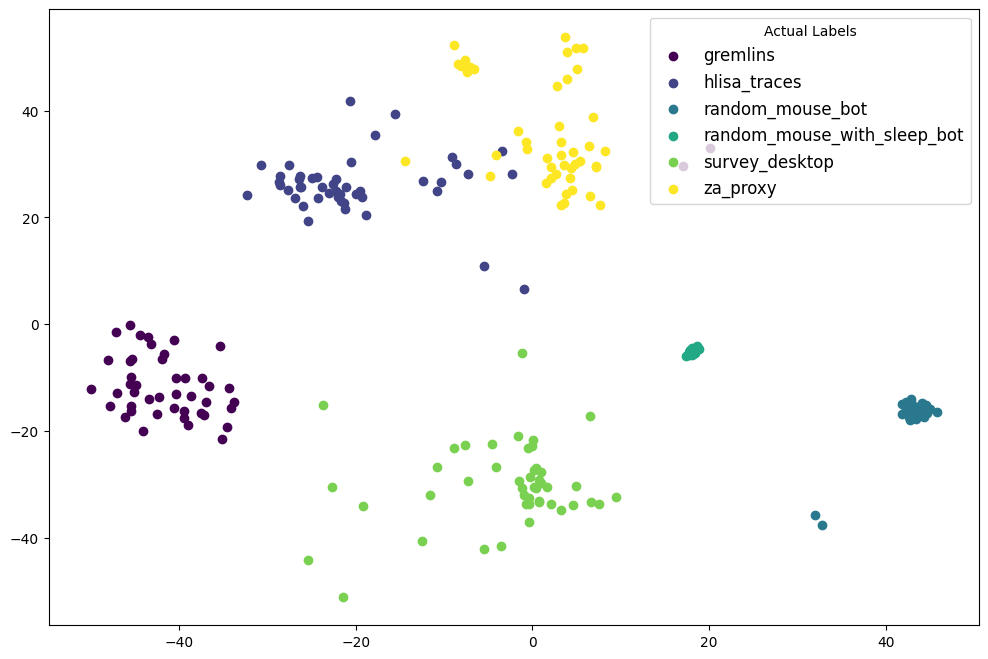}
  \end{minipage}%
  \hfill
  \begin{minipage}{.5\textwidth}
    \centering
    \includegraphics[width=.8\linewidth]{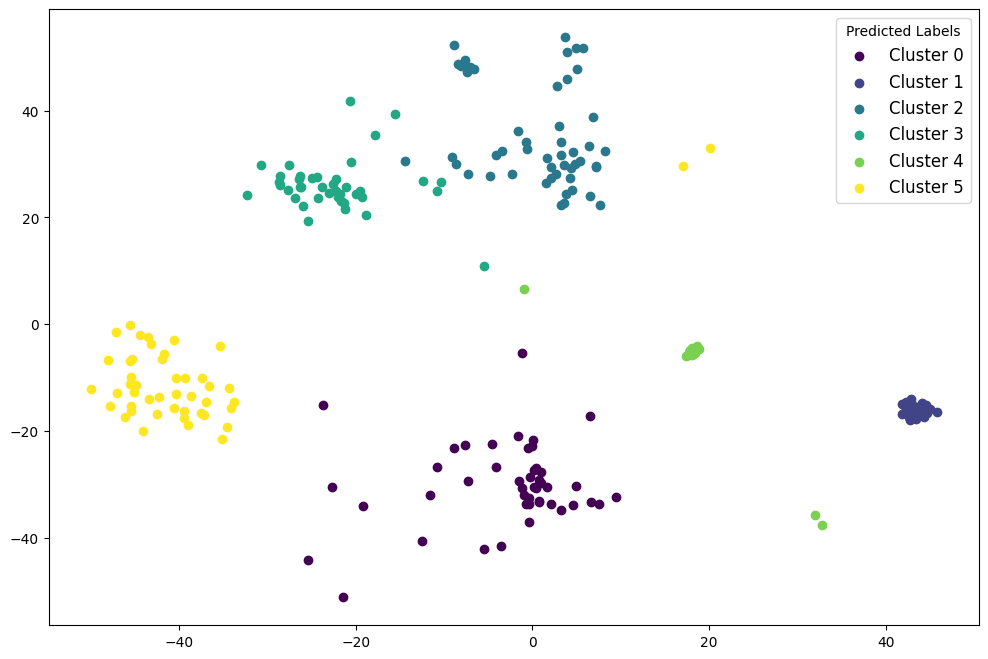}
  \end{minipage}
  \caption{Detecting New Behavioral Trends via Clustering.
  Left:  actual labels. Right: output estimated labels.}
      \label{fig:cl2}
\end{figure*}

\begin{table}[ht]
    \centering
    \caption{Clustering Metrics}
    \label{tab:clustering_metrics}
    \begin{tabular}{l c}
        \toprule
        Metric & Value \\
        \midrule
        Adjusted Rand index & 0.9247 \\
        Adjusted Mutual Information & 0.9234 \\
        Homogeneity & 0.9246 \\
        Completeness & 0.9255 \\
        V-measure & 0.9251 \\
        \bottomrule
    \end{tabular}
\end{table}
To illustrate the fact that \thesystem is capable of detecting new and emerging behavioral trends of automated agents, we use the data from artificially delayed random bots (introduced in Section \ref{sec:artifacts}). The new agent also referred to as `Random Bot with Sleep' is a modified version of the original `Random Bot' with artificial random delays in movements and actions. The resulting output is shown in Figure \ref{fig:cl2} and the metrics are provided in Table \ref{tab:cl2}.


\begin{table}[ht]
    \centering
    \caption{Clustering Metrics with New Trend Introduced}
    \label{tab:cl2}
    \begin{tabular}{l c}
        \toprule
        Metric & Value \\
        \midrule
        Adjusted Rand index & 0.9325 \\
       Adjusted Mutual Information & 0.9394 \\
        Homogeneity & 0.9402 \\
        Completeness & 0.9418 \\
        V-measure & 0.9410 \\
        \bottomrule
    \end{tabular}
\end{table}
\subsection{Classification}
\label{classification}
\begin{figure}[!t]
    \centering
    \includegraphics[width=0.8\linewidth]{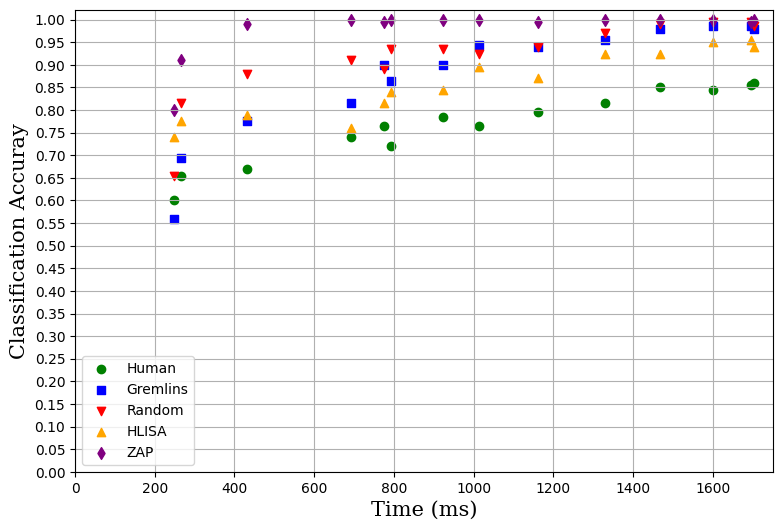}
    \caption{LSTM Accuracy of Classifying Different Agents based on Multi-Modal Artifacts}
    \label{fig:LSTM}
\end{figure}

 For the classification task, we use an LSTM network composed of three successive LSTM layers, each with 300, 200, and 100 units, respectively. These layers utilize the hyperbolic tangent (tanh) activation function. This is followed by a fully-connected dense layer and a final Sigmoid layer. The model was trained on Nvidia Tesla K80 hardware for 50 epochs, completing the training process in a duration denoted as 15 minutes.
During the training phase, each available label was represented by 15,000 artifacts for training purposes, with the remaining artifacts reserved for the test phase. For evaluation, the test iteration parameter was set to 200 per label, as outlined in Algorithm \ref{alg:lstm_classifier}.
%\bo{Reviwer mentioned why didn't use k-fold.}
The test and training phases were performed three times, each with random initialization of trace indices. The reported performance metrics represent the average values obtained from these three runs, providing a comprehensive assessment of the model's generalization and robustness. The accuracy versus time-to-detection tradeoff in multiclass classification is shown in Figure \ref{fig:LSTM}. It can be observed that the probability of misclassification of various automated agents approaches zero in less than one second. 

\subsection{Multi-Modality Analysis}
\label{sec:multi-modal_vs_uni-modal}
\begin{figure}[!t]
    \centering
    \includegraphics[width=0.8\linewidth]{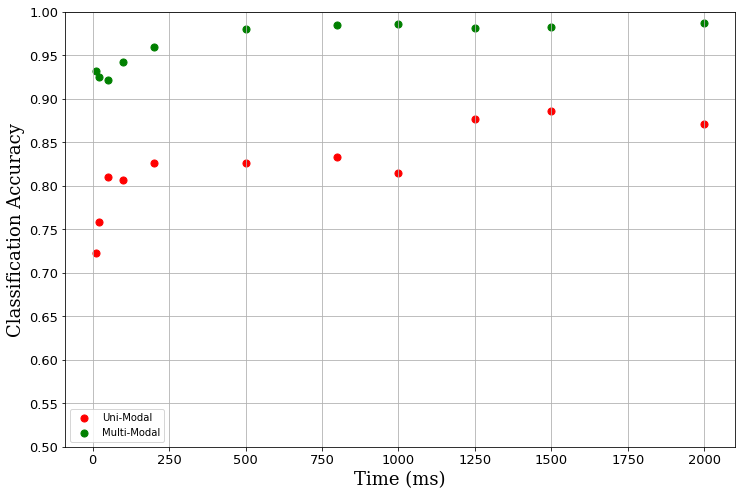}
    \caption{LSTM Classification Uni-Modal vs Multi-Modal}
    \label{fig:LSTM_binary_unimodal_vs_multimodal}
\end{figure}

Prior work has focused on uni-modal behavioral analysis (i.e., mouse movement) for automated traffic detection~\cite{niu2023exploring}. In this section, we empirically analyze the effect of using multi-modal data to uni-modal state-of-the-art approaches.
To this end, we created a new subset of traces and trained two separate models for human-vs-bot classification: one model was trained on the multi-modal data, and the other was trained using the uni-modal. Our analysis shows that the adoption of multi-modal data can impact the accuracy and speed of detection. 
The result of the test phase for the two models is presented in Figure \ref{fig:LSTM_binary_unimodal_vs_multimodal}. In fact, the multi-modal approach achieves on average 8\% higher over the detected time. Furthermore, by continuing the experiment for longer traces, we came to the conclusion that in order to achieve a classification accuracy of 95\% or higher when using
uni-modal data, we are required to use traces with lengths of at
least 10,000 milliseconds which is over 10 times longer for achieving the same results with multi-modal data.
%}
\subsection{Network Overhead Assessment}
\label{sec:overhead}

To assess the network overhead, we conducted experiments that involved embedding the framework's script into a blank HTML webpage. We then manually interacted with the page for a ten-second duration and meticulously recorded every network request. This includes both the payload size and the accompanying request and response headers.
The experiment resulted in the generation of approximately 280 event triggers, with an average payload size of 18KB per second being transmitted to the server. We recorded the sizes for both HTTP and WebSocket-based (Socket) implementations to compare their efficiencies.
\\\textbf{HTTP Overhead:} For the HTTP-based implementation, the average size of request headers was 770 Bytes, and the response headers had an average size of 330 Bytes. These headers introduce redundant overhead for each HTTP request, which when scaled can become significant
\\ \textbf{WebSocket Overhead:} On the other hand, in the WebSocket (Socket) implementation, the request and response headers are sent only once during the connection establishment. Afterward, the recurring overhead is reduced to a meager 10 Bytes for each subsequent payload dispatch.

The empirical data suggests that the WebSocket implementation provides an efficient alternative to HTTP requests, with dramatically reduced recurrent network overhead.

\section{Theoretical Analysis}
\label{sec:th}
In the previous sections, we argued that under the assumptions in \ref{sec:assu}, the classification task can be broken into two steps. First, the defender may try to estimate the probability distribution of activities of humans and various automated agent classes, and then it may use a distance measure such as Jeffreys divergence for classification. Consequently, an attacker trying to evade such defensive mechanisms needs to mimic the statistical distribution of human activities. In Section \ref{sec:multi-modal_vs_uni-modal} we empirically analyzed the effect of using multi-modal data on the classification of the traces promptly.

In this section, we provide theoretical analysis showing that the multi-modal nature of data collection in \thesystem prevents evasion by adversaries in two ways: i) the amount of training data needed to estimate the statistical distribution of human activities, and thus constructing an evading generative model (e.g., via reinforcement learning) grows superlinearly in the number of modalities, and ii) the time-to-detection to achieve a desired probability of false positives and false negatives decreases linearly in the number of modalities. 

\subsection{Effect of Multi-Modal Data Collection on Training Set Complexity}
In this sub-section,   we build upon previous known results to argue that, under the assumption that behavioral biometrics can be modeled as a Markov chain, the amount of training data that an attacker needs to emulate human behavior grows superlinearly with the number of data modalities involved. To elaborate, the error exponent of a hypothesis test applied to Markov sources --- such as a generic human data or adversary's data collected by \thesystem --- is directly proportional to the Kullback-Leibler (KL) divergence between the output distribution of the adversary's generative model and that of a generic human user (e.g., \cite{cover1999elements}). On the other hand, the KL divergence can be bounded using the $\ell_{\infty}$ distance  between the distributions via Pinsker's inequality \cite{csiszar2011information}. 
Furthermore, the $\ell_{\infty}$ distance can be bounded as a function of the number of modalities (i.e., the alphabet size of the Markov process), using the following known result.
\begin{proposition}[\cite{wolfer2021statistical}, Theorem 3.2] 
\label{prop:1}
For every \( \epsilon \in (0, \frac{1}{32}) \), \( \gamma_{\text{ps}} \in (0, \frac{1}{8}) \), \( s = 6k \geq 12 \), and every distribution estimation procedure, there exists an \( s \)-state Markov chain with pseudo-spectral gap \( \gamma_{\text{ps}} \) and stationary distribution \( \pi \) such that the estimation procedure must require a sequence of $m$ samples drawn from the unknown \( M \) of length at least
\[
m \geq c \max \left\{ \frac{s}{\epsilon \sqrt{2\min \pi}}, \frac{s \ln s}{\gamma_{\text{ps}}} \right\},
\]
to achieve $\ell_\infty$ distance less than $\epsilon$, where \( c \) is a universal constant.
\end{proposition}

Consequently, using Proposition \ref{prop:1}, for an adversary such as the one in \cite{akrout1903hacking}, to be able to guarantee an $\ell_{\infty}$ distance less than $\epsilon$ between its output statistics and that of a generic human user, 
the number of necessary training samples  is $\Omega(s\log{s})$, where $s$  is the number of modalities. Furthermore, using Pinsker's inequality \cite{csiszar2011information}, and the fact that variational distance is larger than norm infinity, it follows that in order to achieve a Kullback-Leibler (KL) divergence less than $\epsilon^2$, the attacker requires $\Omega(s\log{s})$ training samples of human data.

\subsection{Effects of Multi-Modal Data on Time-to-Detection}
To rigorously quantify the relationship between time-to-detection and the number of modalities, we consider a simplified scenario involving binary hypothesis testing on sequences of independent and identically distributed (IID) random variables. Let $\mathbf{X}=X_1,X_2,\cdots$ be an IID sequence, taking values from the finite set $\mathcal{X}$. The sequence follows one of two possible distributions: either $P^1_{X}$ or $P^2_X$. The error exponent for the optimal hypothesis test in this setting is  $D_{KL}(P^1_{X}|| P^2_X)$, where $D_{KL}(\cdot||\cdot)$ is the KL divergence. 

To model the uni-modal data, let another sequence $\mathbf{Y}=Y_1,Y_2,\cdots$ be defined as a sampled subsequence from  $\mathbf{X}$, where $Y_j\triangleq X_{i_j}$, $i_1\triangleq \arg\min_i \{i|X_i\in \mathcal{Y}\}$, $i_j\triangleq \arg\min_{i>i_{j-1}} \{i|X_i\in \mathcal{Y}\}$, and $\mathcal{Y}$ is a subset of size two (modalities) chosen randomly and uniformly from the elements of $\mathcal{X}$. 

The following extension of the Chernoff-Stein Lemma \cite[Theorem 11.8.3]{cover1999elements}, shows that the time-to-detection of hypothesis test on uni-modal data compared to that of multi-modal data increases by a factor of $\omega(|\mathcal{X}|)$. 

\begin{lemma}[Sampled Chernoff-Stein Lemma]
\label{lem:SC}
Let $\mathbf{X}$ and $\mathbf{Y}$ be defined as described in the prequel. For  $n\in \mathbb{N}$ and two given hypothesis tests on $\mathbf{X}$ and $\mathbf{Y}$, let $\mathcal{A}_n\in \mathcal{X}^n$ and $\mathcal{A}'_n\in \mathcal{Y}^n$ be  the acceptance regions of the hypothesis $X\sim P^1_X$, respectively. Define the probabilities of error:
\begin{align*}
    &\alpha_n\triangleq P_{X^n}^{1}(\mathcal{A}^c_n),\quad \alpha'_n\triangleq P_{Y^{N_n}}^{1}(\mathcal{A}'^c_{N_n}),
    \\&\beta_n\triangleq P_{X^n}^{2}(\mathcal{A}_n),\quad \beta'_n\triangleq P_{Y^{N_n}}^{2}(\mathcal{A}'_{N_n}),
\end{align*}
where $N_n\triangleq |\{i|X_i\in \mathcal{Y}, i\in [n]\}|$. For $\epsilon\in [0,\frac{1}{2}]$, define:
\begin{align*}
    &\beta^{\epsilon}_n\triangleq \min_{\mathcal{A}_n\subseteq \mathcal{X}^n, \alpha_n\leq \epsilon} \beta_n, 
    \quad {\beta'}^{\epsilon}_n\triangleq \min_{\mathcal{A}'_n\subseteq \mathcal{Y}^n, \alpha'_n\leq \epsilon} \beta'_n.
\end{align*}
Then, 
\begin{align*}
    \lim_{n\to \infty} \frac{1}{n}\log{\beta^{\epsilon}_n}=\omega(|\mathcal{X}|) \lim_{n\to \infty} \frac{1}{n}\log{\beta'^{\epsilon}_n}.
\end{align*}
\end{lemma}
The proof is provided in the following.

\subsection{Proof of Lemma \ref{lem:SC}}
\label{app:lem:SC}
\begin{proof}
    We provide an outline of the proof. Note that the error exponent of the hypothesis test on $\mathbf{Y}$ is given by:
    \begin{align*}
        \frac{1}{n}\mathbb{E}_{X^n}(\log{\frac{P^1_{Y^{N_n}}(Y^{N_n})}{P^2_{Y^{N_n}}(Y^{N_n})}}),
    \end{align*}
    where $X^n$ is produced based on $P_X^1$, and $P^1_{Y^n}$ and $P^2_{Y^n}$ are distributions induced by $P_X^1$ and $P_X^2$, respectively. Using the IID property of $\mathbf{Y}$, and the fact that $N_n$ is renewal process, Wald's identity yields:
    \begin{align*}
        \mathbb{E}_{X^n}(\log{\frac{P^1_{Y^{N_n}}(Y^{N_n})}{P^2_{Y^{N_n}}(Y^{N_n})}})= \mathbb{E}(N_n) \mathbb{E}(\log\frac{P^1_Y(Y)}{P^2_Y(Y)}).
    \end{align*}
    We have $\mathbb{E}(N_n)=nP(X\in \mathcal{Y})$. Also, $P_Y^i(y)= P_X^i(y|X\in \mathcal{Y})= \frac{P_X^i(y)}{P^i_X(\mathcal{Y})}, i\in \{1,2\}, y\in \mathcal{Y}$. So, 
    \begin{align*}
      \mathbb{E}_{X^n}(\log{\frac{P^1_{Y^{N_n}}(Y^{N_n})}{P^2_{Y^{N_n}}(Y^{N_n})}})
      &=\mathbb{E}(\sum_{y\in \mathcal{Y}}P^1_X(y)\log{\frac{P^1_X(y)}{P^2_X(y)}})+
      \\&P^1(X\in \mathcal{Y})\log\frac{P^2(X\in \mathcal{Y})}{P^1(X\in \mathcal{Y})}
    \end{align*}
    It follows from the assumption that $\mathcal{Y}$ is a subset chosen randomly and uniformly from $\mathcal{X}$ that 
    \begin{align*}
       \mathbb{E}(\sum_{y\in \mathcal{Y}}P^1_X(y)\log{\frac{P^1_X(y)}{P^2_X(y)}})
       = \frac{|\mathcal{Y}|}{|\mathcal{X}|}\mathbb{E}(\log{P^1_X(X)}{P^2_X(X)}).
    \end{align*}
    This concludes the proof. 
\end{proof}

The key implication of Lemma \ref{lem:SC} is that to achieve a comparable error exponent, the time-interval for observing uni-modal data must be scaled by a factor of $\frac{|\mathcal{X}|}{|\mathcal{Y}|}$, which is linearly proportional to the number of modalities. Therefore, we have shown that, under the IID input assumption, the time-to-detection can be linearly reduced by incorporating an increasing number of modalities.

%% file: 06-Discussion.tex
\section{Discussion}
\label{Discussion}
%\subsection{Key Takeaways}
\noindent \textbf{Enhanced Visibility and Minimal Changes:} One of the central claims of this paper is the enhancement in visibility into web scanning activities, achievable with minimal to no alterations to the target web applications or the user experience. We introduce a lightweight, multi-modal forensics engine designed to extend the capabilities of existing security systems. By applying the forensics engine and the associated machine learning techniques to real-world data, and also by providing theoretical analysis, we demonstrate that the acquisition of multi-modal data enables robust defense mechanisms. These mechanisms not only effectively differentiate between human users and various classes of automated web scanners but also do so in an unsupervised manner.

\noindent \textbf{Elevated Evasion Costs:} Another advantage of \thesystem is its impact on the economics of evasion studied in Section \ref{sec:th}. By utilizing multi-modal data types, our approach increases the complexity and cost associated with evading detection and shortens the detection time compared to uni-modal monitoring systems. 
This has the dual benefit of deterring less sophisticated attackers while providing a stronger defense against advanced threats. 

\noindent \textbf{Strategic Importance in a Complex Environment:} In the web ecosystem, where adversaries are continuously enhancing 
their attack payloads to operate faster and cheaper, techniques similar to \thesystem 
%new security incidents  is an intricate and continually evolving battleground for security professionals. In this complex environment, innovative security systems like 
%\thesystem 
can serve as vital tools in tilting the defense asymmetry in favor of defenders. 
Not only do they enhance the ability to detect and classify threats, but they also contribute to a richer understanding of the broader attack landscape.
\thesystem is agnostic to web technologies being used in target web applications and introduce minimal integration cost to the target systems, offering flexibility for wide deployment in the modern web where different underlying technologies are being used in practice. 
% \noindent \textbf{More Visibility.} In this paper, we claim that the visibility over web scanning activities can be enhanced significantly with almost no changes to the target web applications or user experience. We proposed a light-weight multi-modal forensics engine and showed that with access to multi-modal data, it possible to build robust defense mechanisms to differentiate different web scanners from human users in an unsupervised fashion while increasing the cost of evasion. 
% In a complex and dynamic environment such as web, security systems such as \thesystem are an important step towards improving the defense asymmetry against adversarial scanners and achieving better understanding of the attack landscape.

%The proposed approach is agnostic to web technologies being used in modern web ecosystem and does not introduce any significant integration cost to the target systems. We showed that with multi-modal artifacts, similar to \thesystem's, 

\noindent \textbf{Resistance Against Behavioral Perturbation:} One of the major hurdles in effective bot detection is countering the flexibility that attackers possess in altering their behavior patterns. This issue, termed as incomplete mediation~\cite{cisa_2005}, arises when adversaries exploit vulnerabilities, loopholes, and corner cases to avoid detection by diverging from anticipated behavior norms. According to our theoretical results, the proposed multi-modal forensics engine successfully mitigates this challenge by escalating the cost of behavioral perturbation for the attackers linearly as a function of the number of monitored modalities. This results in a more resilient and robust defense mechanism against such evasion tactics. However, it is critical to acknowledge that while these measures enhance the efficacy of the defense system, they are not a panacea. With the continual advancements in generative machine learning techniques, there remain pressing questions regarding the theoretical feasibility of bot detection when facing adversaries with virtually unlimited resources. These concerns have recently received attention in works that explore detection challenges in broader contexts, e.g., ~\cite{sadasivan2023can}.

% \ak{we should be careful about preserving privacy claim...We should somewhere say we achive the results with zero knowledge on PII info, such as ip address.}

\noindent \textbf{Identification and Behavioral Cataloging.} We showed that low-latency traffic attribution can be done effectively %with access to the multi-modal artifacts.
%In addition to detecting non-human traffic, perhaps the more important feature of \thesystem is to build a robust mechanism to reason about the dynamic behavior of the web session and detect previously unknown web sessions without reliance on potentially identifiable information such as IP addresses and fingerprinting data. The proposed attribution mechanism can be used to generate reliable signature of dynamic behavior and build a large catalog of automated threats seen in the wild.
%In particular, we showed that the attribution module can offer insights on the dynamic behavior of the web sessions and identify even previously unknown web sessions as a new observation or data drift.
%The attribution and cataloging mechanism functions 
with almost zero access to labeled data or potentially identifiable information such as IP addresses or fingerprintable data. We also showed that the multi-modal artifacts generated by \thesystem can be used to build a real-time detection mechanism, differentiating real user traffic from non-human web traffic with varied levels of complexity in less than one second with an accuracy of 96\%.
These results were achieved 
%That is, \thesystem operates
without any prior assumption on the type and volume of web traffic, underscoring its suitability for wide adoption.

%\noindent \textbf{Low-Latency Detection} In addition to behavioral cataloging, we also showed that the multi-modal artifacts generated by \thesystem can be used to build a real-time bot detection mechanism, differentiating real user traffic from non-human web traffic with varied level of complexity in less than 250ms with the accuracy of 96\%. We used traces from 5 different types of bots together with the data from nearly 50 human sessions to train the LSTM model. In Section~\ref{sec:impacts}, we elaborate on how the module can be used.

%\subsection{Impacts on the state-of-the art}
\label{sec:impacts}

%\noindent \textbf{Work Factor.} The design of \thesystem is in line with the concept of economy of mechanism design principle~\cite{cisa_2005}. That is, the proposed defense should introduce low implementation and deployment cost on the defense side while adding significant defense asymmetry. 
%as it introduces very low overhead in implementing and deploying the collective approach on the defense side, but adds significant defense asymmetry.
%To enable web applications with \thesystem, it only takes one javascript tag to insert into the web application without requiring any changes to the architecture or underlying logics of the program. At the same time, it increases the cost of evasion linearly as shown in section x.

%\ak{farhad can you confirm/extend the last part on increasing the cost based on the content you added?}
\noindent \textbf{Integration into Current Defenses.} Services such as Google Analytics~\cite{googleAnalytics}, DataDog~\cite{datadog}, HotJar~\cite{hotjar}, and MonsterInsights~\cite{monsterinsights} --- which provide software supply chain management or analytical insights about page views, page ranking, and audience insight ---  are used in millions of high-profile websites. We envision that defensive monitoring engines such as \thesystem can easily augment such services to enable offensive scan detection, analysis, and blocklisting with minimal changes to the code base, enabling millions of websites with a low-overhead defense mechanism.

%% file: 07-Conclusion.tex
\section{Conclusions}
\label{Conclusions}
In conclusion, the rising complexities and financial implications of automated web traffic demand robust solutions. We introduced \thesystem, an in-application forensics engine designed to meet key objectives: seamless integration across different platforms, minimal overhead, real-time detection, and offline attribution of automated agents. Leveraging multi-modal behavioral monitoring and machine learning architectures, \thesystem achieves high levels of detection and attribution accuracy. Our theoretical and empirical evaluations affirm its effectiveness, showing above 90\% accuracy in detection within hundreds of milliseconds and minimal communication overhead. Theoretical and empirical analysis was provided to illustrate the effectiveness of the proposed defense mechanisms. 
\balance
%Web bots tirelessly explore the web, fulfilling diverse functions ranging from search engine indexing and data scraping to website monitoring and security checks. On the flip side, malicious bots engage in activities like web scraping for sensitive information or launching distributed denial-of-service (DDoS) attacks. In this paper, we presented a multi-modal forensics engine called \thesystem for bot detection and classification. The proposed framework provides the required artifacts for online bot detection as well as offline attribution which can be used to classify different types of bots. Additionally, we tested the framework's ability to identify the emergence of new behavior by injecting artificially manufactured mouse movements.
%We provided theoretical and empirical analysis of how multiple modalities can help minimize time to detection and make it harder for automated agents to generate traces similar to humans. Finally, we analyze the network overhead and discuss how the framework can be incorporated into the currently available solutions.

%% file: 09-Apendix.tex
% \newpage
% \clearpage
% \vfill\break
\appendix
\section{Comprehensive List of Monitored Events}
\label{appendix:monitored_events}

In the interest of providing a detailed reference for readers, this appendix enumerates all the Document and Window Events that are monitored and logged by \thesystem. These events serve as the foundational data points for both human and automated interactions with the web application. Each event is categorized based on its domain of activity, either as a Document Event or a Window Event, as specified in Tables~\ref{tab:events}.

\section{Proof of Lemma \ref{lem:SC}}
\label{app:lem:SC}
\begin{proof}
    We provide an outline of the proof. Note that the error exponent of the hypothesis test on $\mathbf{Y}$ is given by:
    \begin{align*}
        \frac{1}{n}\mathbb{E}_{X^n}(\log{\frac{P^1_{Y^{N_n}}(Y^{N_n})}{P^2_{Y^{N_n}}(Y^{N_n})}}),
    \end{align*}
    where $X^n$ is produced based on $P_X^1$, and $P^1_{Y^n}$ and $P^2_{Y^n}$ are distributions induced by $P_X^1$ and $P_X^2$, respectively. Using the IID property of $\mathbf{Y}$, and the fact that $N_n$ is renewal process, Wald's identity yields:
    \begin{align*}
        \mathbb{E}_{X^n}(\log{\frac{P^1_{Y^{N_n}}(Y^{N_n})}{P^2_{Y^{N_n}}(Y^{N_n})}})= \mathbb{E}(N_n) \mathbb{E}(\log\frac{P^1_Y(Y)}{P^2_Y(Y)}).
    \end{align*}
    We have $\mathbb{E}(N_n)=nP(X\in \mathcal{Y})$. Also, $P_Y^i(y)= P_X^i(y|X\in \mathcal{Y})= \frac{P_X^i(y)}{P^i_X(\mathcal{Y})}, i\in \{1,2\}, y\in \mathcal{Y}$. So, 
    \begin{align*}
      \mathbb{E}_{X^n}(\log{\frac{P^1_{Y^{N_n}}(Y^{N_n})}{P^2_{Y^{N_n}}(Y^{N_n})}})
      &=\mathbb{E}(\sum_{y\in \mathcal{Y}}P^1_X(y)\log{\frac{P^1_X(y)}{P^2_X(y)}})+
      \\&P^1(X\in \mathcal{Y})\log\frac{P^2(X\in \mathcal{Y})}{P^1(X\in \mathcal{Y})}
    \end{align*}
    It follows from the assumption that $\mathcal{Y}$ is a subset chosen randomly and uniformly from $\mathcal{X}$ that 
    \begin{align*}
       \mathbb{E}(\sum_{y\in \mathcal{Y}}P^1_X(y)\log{\frac{P^1_X(y)}{P^2_X(y)}})
       = \frac{|\mathcal{Y}|}{|\mathcal{X}|}\mathbb{E}(\log{P^1_X(X)}{P^2_X(X)}).
    \end{align*}
    This concludes the proof. 
\end{proof}

% \begin{table*}[!h]
% \centering
% \caption{Document and Window Events}
% \label{tab:events}
% \begin{tabular}{|c|l|c|l|c|l|c|l|}
% \hline
% \multicolumn{2}{|c|}{Document Events} & \multicolumn{2}{c|}{Document Events (Cont.)} & \multicolumn{2}{c|}{Window Events} & \multicolumn{2}{c|}{Window Events (Cont.)}\\
% \hline
% Index & Event & Index & Event & Index & Event & Index & Event \\
% \hline
% 0 & mousedown & 13 & keypress & 25 & load & 38 & pagehide \\
% 1 & mouseup & 14 & click  & 26 & unload & 39 & pageshow \\
% 2 & mousemove & 15 & dblclick & 27 & beforeunload & 40 & message \\
% 3 & mouseover & 16 & scroll  & 28 & blur & 41 & beforeprint \\
% 4 & mouseout & 17 & change & 29 & focus & 42 & afterprint \\
% 5 & mousewheel & 18 & select & 30 & resize &  &  \\
% 6 & wheel & 19 & submit  & 31 & error &  &  \\
% 7 & touchstart & 20 & reset & 32 & abort &  &  \\
% 8 & touchend & 21 & context menu & 33 & online &  &  \\
% 9 & touchmove & 22 & cut & 34 & offline &  &  \\
% 10 & deviceorientation & 23 & copy & 35 & storage &  &  \\
% 11 & keydown & 24 & paste  & 36 & popstate &  &  \\
% 12 & keyup &  &  & 37 & hashchange &  &  \\
% \hline
% \end{tabular}
% \end{table*}

\begin{table}[h]
\centering
\caption{Document and Window Events}
\label{tab:events}
\begin{tabular}{|c|l|c|l|}
\hline
\multicolumn{4}{|c|}{Document Events} \\
\hline
Index & Event & Index & Event \\
\hline
0 & mousedown & 13 & keypress \\
1 & mouseup & 14 & click \\
2 & mousemove & 15 & dblclick \\
3 & mouseover & 16 & scroll \\
4 & mouseout & 17 & change \\
5 & mousewheel & 18 & select \\
6 & wheel & 19 & submit \\
7 & touchstart & 20 & reset \\
8 & touchend & 21 & context menu \\
9 & touchmove & 22 & cut \\
10 & deviceorientation & 23 & copy \\
11 & keydown & 24 & paste \\
12 & keyup & & \\
\hline
\multicolumn{4}{c}{} \\  % Add some space between the two sets of events
\hline
\multicolumn{4}{|c|}{Window Events} \\
\hline
Index & Event & Index & Event \\
\hline
25 & load & 37 & hashchange\\
26 & unload &36 & popstate  \\
27 & beforeunload &35 & storage \\
28 & blur & 34 & offline \\
29 & focus & 38 & pagehide  \\
30 & resize & 39 & pageshow \\
31 & error & 40 & message\\
32 & abort & 41 & beforeprint \\
33 & online & 42 & afterprint \\
\hline
\end{tabular}
\end{table}
% \begin{figure}[!h]
%     \centering
%     \includegraphics[width=0.9\linewidth]{images/lstm_heatmaps.png}
%     \caption{LSTM Heat Map for Different Thresholds}
%     \label{fig:lstm_heatmap}
% \end{figure}

\begin{figure}[!t]
    \centering
    \includegraphics[width=\linewidth]{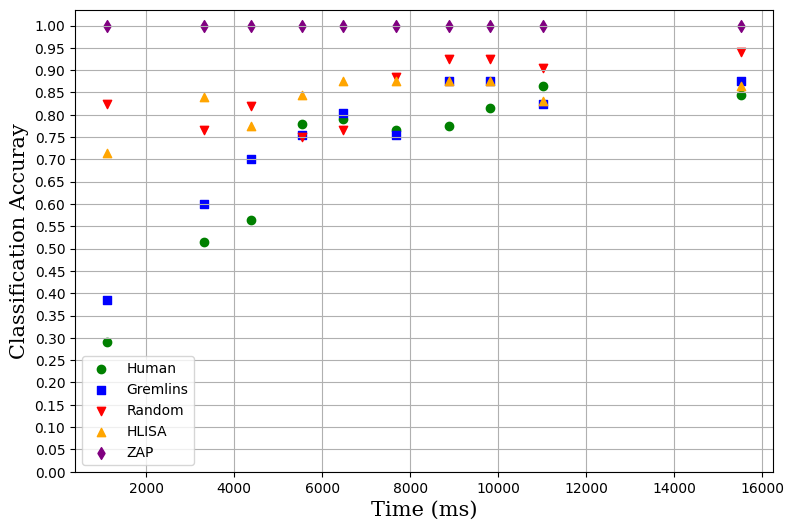}
    \caption{LSTM Accuracy of Classifying Different Agents based on Uni-Modal Artifacts (Mousemove Only)}
    \label{fig:LSTM_unimodal}
\end{figure}
\section{LSTM Additional Experiments}
As mentioned in Section \ref{classification}, we have conducted several additional experiments using the LSTM model. In particular, in order to study the effect of multi-modal artifacts, we designed a separate experiment to use only the mouse movement event. The result of the accuracy of the LSTM model trained for this task is presented in Figure \ref{fig:LSTM_unimodal}. 
% The heatmap and time-to-detections are also provided for various hyperparameter choices in Figure \ref{fig:lstm_heatmap}.

% \begin{figure}[!h]
%     \centering \includegraphics[width=0.9\linewidth]{images/lstm_single_modality_5labels.png}
%     \caption{LSTM Uni-Modal Classification Accuracy (Mousemove Only)}
%     \label{fig:LSTM2}
% \end{figure}

\begin{figure}[!h]
    \centering
    \includegraphics[width=\linewidth]{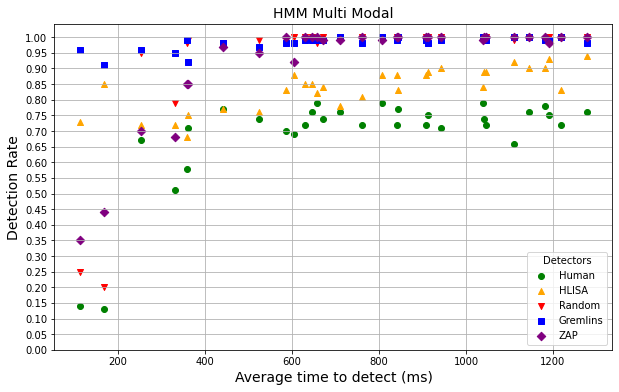}
    \caption{HMM Accuracy of Classifying Different Agents based on Multi-Modal Artifacts}
    \label{fig:HMMMultiModal}
\end{figure}

% \noindent \textbf{Binary Classification}. One of the use cases of the proposed framework is to successfully detect humans against bots. Therefore, we designed an experiment with only the two classes. Figure \ref{fig:LSTM_bin} illustrates the accuracy of the model for binary classification task.
% \begin{figure}[h]
%     \centering
%     \includegraphics[width=\linewidth]{images/lstm_binary_acc_10ts.png}
%     \caption{LSTM Accuracy for Bot vs Human Classification}
%     \label{fig:LSTM_bin}
% \end{figure}

% \begin{figure}[h]
%     \centering
%     \includegraphics[width=\linewidth]{images/HMMBinary.png}
%     \caption{HMM Accuracy for Bot vs Human Classification}
%     \label{fig:HMM_bin}
% \end{figure}

% \begin{figure}[!t]
%     \centering
%     \includegraphics[width=0.9\linewidth]{images/hmm_heatmaps.png}
%     \caption{HMM Heat Map for Different Thresholds}
%     \label{fig:HMMHeat}
% \end{figure}

\section{HMM Classification Results}
\label{A:hmm:classification}
Simulations for the classification in section \ref{classification} were repeated with the HMM model. During the training phase, each available label was represented by 15,000 artifacts for training purposes, with the remaining artifacts reserved for the test phase.  
% Furthermore, 
% the heatmap (Figure \ref{fig:HMMHeat}) presents a more granular view of the classification results under different threshold settings.
% Four distinct heat maps represent classification outcomes at thresholds of 0, 10, 20, and 30. The color intensities depict the classification counts, with deeper shades indicating higher counts. True classes are represented on the y-axis while the predicted classes are on the x-axis. It’s evident from these heatmaps that increased threshold values leads to improved accuracy with accuracy of above \%90 achieved in under a second of monitoring. Additionally, 
Figure \ref{fig:HMMMultiModal} illustrates the tradeoff between accuracy and time-to-detection in multi-class classification, indicating the average duration each detector requires to identify a bot.